\documentclass[12pt]{article}
\usepackage{mathtools,amsthm}
\usepackage{graphicx}
\usepackage{enumerate}
\usepackage{natbib}
\usepackage{url}
\usepackage[hidelinks]{hyperref}
\usepackage{bibunits}

\usepackage{times}
\usepackage{amsbsy}
\usepackage{amssymb}
\usepackage{mathrsfs}
\usepackage{bm}
\usepackage{enumerate}
\usepackage[ruled,linesnumbered]{algorithm2e}
\usepackage{color}
\usepackage{xcolor}
\usepackage{authblk}
\newtheorem{theorem}{Theorem}
\newtheorem{lemma}{Lemma}
\newtheorem{condition}{Condition}
\newtheorem{corollary}{Corollary}
\newtheorem{remark}{Remark}
\newtheorem{definition}{Definition}

\newcommand{\tildeY} {\widetilde{Y}}

\newcommand{\e}{\varepsilon}
\newcommand{\ei}{\varepsilon_i}
\newcommand{\tildePsi}{\widetilde{\Psi}^{(i)}_S}
\newcommand{\tildePsij}{\widetilde{\Psi}^{(i)}_k}
\newcommand{\tildePsijT}{\widetilde{\Psi}^{(i)\T}_k}
\newcommand{\tildePsiT}{\widetilde{\Psi}^{(i)\T}_S}
\newcommand{\tildePsino}{\widetilde{\Psi}^{(-i)}_S}
\newcommand{\tildePsinoT}{\widetilde{\Psi}^{(-i)T}_S}

\newcommand{\betaStno}{\beta_S^{*(-i)}}
\newcommand{\betaSti}{\beta_S^{*(i)}}

\newcommand{\hatgSi}{\hat{g}_S^{(i)}}
\newcommand{\hatgSCi}{\hat{g}_{S^c}^{(i)}}
\newcommand{\hatgji}{\hat{g}_j^{(i)}}
\newcommand{\hatgjiT}{\hat{g}_j^{(i)\T}}
\newcommand{\onen}{\frac{1}{n}}

\newcommand{\SigmaSS}{\Sigma_{SS}^{(i)}}

\newcommand{\SigmaSSinv}{\Sigma_{SS}^{(i)-1}}

\newcommand{\lef}{\left\|}
\newcommand{\rig}{\right\|}
\newcommand{\sumjp}{\sum_{j=1}^{p}}
\newcommand{\rP}{\mathrm{P}}

\DeclareMathOperator{\tr}{tr}

\def\T{{ \mathrm{\scriptscriptstyle T} }}

\newenvironment{myproof}[1][\proofname]{
  \proof[\bfseries \upshape #1]
}{\endproof}

\newcommand{\blind}{1}

\date{}
\begin{document}

\defaultbibliographystyle{agsm}
\defaultbibliography{boyd2011distributed}

\def\spacingset#1{\renewcommand{\baselinestretch}%
  {#1}\small\normalsize} \spacingset{1}

\if1\blind
{\title{\bf DDAC-SpAM: A Distributed Algorithm for Fitting High-dimensional Sparse Additive Models with Feature Division and Decorrelation \let\thefootnote\relax\footnotetext{He and Wu contribute equally to this work. }
}
  \author[1]{Yifan He}
  \affil[1]{Department of Statistics, The Chinese University of Hong Kong}
  \author[2]{Ruiyang Wu}
  \affil[2]{Department of Biostatistics, School of Global Public Health\\New York University}
  \author[3]{Yong Zhou}
  \affil[3]{Academy of Statistics and Interdisciplinary Sciences and School of Statistics, East China Normal University}
  \author[2]{Yang Feng}
  \maketitle
} \fi

\bigskip
\begin{abstract}
Distributed statistical learning has become a popular technique for large-scale data analysis. Most existing work in this area focuses on dividing the observations, but we propose a new algorithm, DDAC-SpAM, which divides the features under a high-dimensional sparse additive model. Our approach involves three steps: divide, decorrelate, and conquer. The decorrelation operation enables each local estimator to recover the sparsity pattern for each additive component without imposing strict constraints on the correlation structure among variables. The effectiveness and efficiency of the proposed algorithm are demonstrated through theoretical analysis and empirical results on both synthetic and real data. The theoretical results include both the consistent sparsity pattern recovery as well as statistical inference for each additive functional component. Our approach provides a practical solution for fitting sparse additive models, with promising applications in a wide range of domains.
\end{abstract}

\noindent%
{\it Keywords:}  Divide, decorrelate and conquer; Feature space partition; Consistency;  Variable selection; Additive model.
\vfill

\begin{bibunit}
\newpage
\spacingset{1.9} %
\section{Introduction}
\label{sec1}
In modern statistics and computing practices, there exists a common
bottleneck that complex data with unprecedented size cannot fit into
memory nor be analyzed within reasonable time on a single machine.
One popular solution is distributed statistical learning which works
by distributing the learning task to different machines and combining
the estimates afterward
\citep{boyd2011distributed,zhang2012communication,
  lee2017communication}. According to the way of partitioning the
dataset, we call a method \emph{observation-distributed} or
\emph{feature-distributed}.  Substantial progress has been made on the
former type that partitions observations into subsets and then fits
the same model using each subset with the same features in different
machines. Most of the literature focuses on the massive linear or
generalized linear models \citep{chen2014split, battey2015distributed,
  zhang2015divide, zeng2015random, tang2016method, zhao2016partially,
  he2016sparse, lee2017communication, shi2018massive}.  For
nonparametric inference, the existing studies include a partial linear
model \citep{zhao2016partially} and nonparametric regression model
\citep{zhang2013divide}.  However, to the best of our knowledge, the
feature-distributed statistical learning method, especially for high
dimensional nonparametric regression, still remains to be developed.

In this paper, we propose a \emph{feature-distributed} algorithm for sparse additive model
with potentially massive number of covariates ($p\gg n$). This problem can be formulated as follows: we are given $n$ observations with response $y_i \in \mathbb{R}$ and covariates $\{x_{i1},\ldots,x_{ip}\} \in \mathbb{R}^p$ for $i=1,\ldots, n$. The goal is to fit the additive model \citep{stone1985additive}
\begin{equation}
  \label{1}
  y_i=\sumjp f_j(x_{ij})+\ei,
\end{equation}
where the number of covariates $p$ can grow much faster than the
sample size $n$ with $\log(p)=n^{v}$ for some $v\in (0,1)$.  What
makes the high dimensional inference possible is the sparsity
assumption where only a small subset of $\{f_j, j=1,\ldots,p\}$ are
nonzero functions.  Many sparsity-promoted estimators have been
proposed for (\ref{1})  \citep{aerts2002some, ravikumar2009sparse,
  meier2009high, koltchinskii2010sparsity, huang2010variable,
  raskutti2012minimax, yuan2016minimax, petersen2016fused,
  sadhanala2017additive}. Since sparse additive models (SpAM) are
essentially a functional version of the group lasso,
\cite{ravikumar2009sparse} borrowed ideas from the sparse linear model
and proposed a corresponding algorithm for solving the problem
\begin{equation*}
  \min_{\beta_j\in \mathbb{R}^{d_n},  j=1,\ldots, p}\frac{1}{2n}\lef Y-\sumjp \Psi_j\beta_j \rig_2^2 +\lambda\sumjp\sqrt{\onen \beta_j^{\T}\Psi_j^{\T}\Psi_j\beta_j}.
\end{equation*}
where $\beta_j=(\beta_{j1}, \ldots, \beta_{jd_n})^{\T}$ is a length-$d_n$ vector of coefficients, and $\Psi_j=(\psi_{j1},\ldots, \psi_{jd_n})$ is an $n\times d_n$ matrix of the truncated set of orthogonal basis functions for $f_j$ evaluated at the training data.
Minimax optimal rates of convergence were established in \cite{raskutti2012minimax} and \cite{yuan2016minimax}. Other extensions of the high dimensional additive model have been proposed by \cite{lou2016sparse, sadhanala2017additive, petersen2019data} and \cite{ Haris2019Generalized}.

Considering datasets of massive dimensions that the computational
complexity or memory requirements cannot fit into one single computer,
the aforementioned frameworks for additive models are not directly
applicable.  A distributed solution and the decentralized storage of
datasets are necessary.  Most works on distributed statistical
inference assume that the data is partitioned by observations, because
of the good theoretical properties of the averaged
estimators. However, under the high-dimensional additive model, the
spline basis functions are constructed from the whole sample. Besides,
each component is represented by a group of basis functions which
results in an even higher dimension of the design matrix than the
number of observations. It is thus desirable to seek
feature-distributed algorithms. But feature-distributed studies are
scarce, partially due to the fact that the feature distribution
process which ignores the correlation between covariates could lead to
incorrect inference. Indeed, dividing feature space directly usually
leads to misspecified models and ineradicable bias.

Next, we review several works on feature distributed methods.
Inspired by \emph{group testing}, \cite{zhou2014parallel} proposed a
\emph{parallelizable feature selection} algorithm. They randomly
sectionalized features repeatedly for tests and then ranked the
features by the test scores.  This attempt can boost efficiency but
its success heavily depends on the correlation structure of
covariates. \cite{song2015split} proposed a Bayesian variable
selection approach for ultrahigh dimensional linear regression models
based on splitting feature set into lower-dimensional subsets and
screening important variables respectively with the marginal inclusion
probability for final aggregation. Similar treatments can be found in
the \cite{yang2016feature}, although in the final stage they utilized
the \emph{sketch} approach for further selection. The efficiency of
this kind of algorithms will again be highly affected by the
correlation structure among features.  Thus the identifiability
condition for controlling the degree of multicollinearity is
necessary.  Based on those key facts, \cite{wang2016decorrelated}
relaxed the correlation requirements by preprocessing the data with a
\textsl{decorrelation} operator (DECO) to lower the correlation in
feature space under the linear regression model. In a related work,
this decorrelation operator was shown to satisfy the irrepresentable
condition for \textsl{lasso} \citep{jia2015preconditioning}. With
DECO, we can get consistent estimates of coefficients with
misspecified submodels.

In this work, we consider decorrelating covariates and propose the feature-distributed algorithm  DDAC-SpAM under the high-dimensional additive model. That is, we first divide the whole dataset by predictors, i.e. each local machine operates on only $p_i$ variables. Then local machines approximate each component in additive models with a truncated set of B-spline basis. After decorrelating the design matrix of the B-spline basis with the central machine, local machines can in parallel conduct group lasso fit efficiently. Finally, the central machine combines the discovered important predictors and refines the estimates. This algorithm can be regarded as a functional extension of the DECO procedure proposed by \cite{wang2016decorrelated}. It provides an efficient way of conducting simultaneous feature selection and point estimation. On top of that, we incorporate a debiasing step \citep{cai2022sparse,van2014asymptotically} and propose a chi-squared test for each functional summands in model \eqref{1}.

The rest of this article is organized as follows. In Section
\ref{sec2}, we review the sparse additive model problem. In Section
\ref{sec3}, we introduce the distributed feature selection procedure
for the additive model after decorrelation. In Section
\ref{sec:stat-infer-via}, we construct a chi-squared test based on a debiased version of the
DDAC-SpAM algorithm. In Section \ref{sec4}, we present the
\emph{sparsistency} property (i.e. sparsity pattern consistency)
\citep{ravikumar2009sparse} of the DDAC-SpAM algorithm and asymptotic
theories for the hypothesis testing framework. Our simulations and a
real data analysis are presented in Sections \ref{sec5} and \ref{sec6}, respectively,
showing the efficiency and effectiveness of our method. We conclude
with a discussion in Section \ref{sec7}. All the technical details are
relegated to the supplementary material.

Some standard notation used throughout this paper is collected
here. For number $a$, $\lceil a\rceil$ represent the smallest integer
larger than or equal to $a$. For a square matrix $A$, let
$\lambda_{\textrm{min}}(A)$, $\lambda_{\textrm{max}}(A)$ and
\(\tr(A)\) denote the minimum and maximum eigenvalues and the
trace. We use the norms $\|A\|=\sqrt{\lambda_\textrm{max}(A^{\T} A)}$,
\(\|A\|_F = \sqrt{\tr(A^{\T} A)}\) and
$\|A\|_{\infty}=\max_{i}\sum_{j=1}^{n}|A_{ij}|$. For vector
$v=(v_1,\ldots,v_k)^{\T}$, we use the norms
$\|v\|=\sqrt{\sum_{j=1}^{k}v_j^2}$ and
$\|v\|_\infty=\max_{j}|v_j|$. For function \(f\), \(f = 0\) means
\(f\) is the zero constant function.

\section{Sparse Additive Model}
\label{sec2}
Given a random sample $\{(x_{i1}, \ldots, x_{ip}), y_i\}_{i=1}^n$, where for each $j$, 
 \(\{x_{ij}, i = 1, \ldots, n\} \stackrel{\rm{iid}}{\sim} \mu_j\) in which \(\mu_j\) is a
probability distribution supported on \([0, 1]\), we consider the nonparametric
additive model
\begin{align*}
  y_i=\sumjp f_j(x_{ij})+\varepsilon_i,
\end{align*}
where the error
$\varepsilon_i \stackrel{i.i.d.}{\sim} \mathcal{N}(0,\sigma^2)$,
$i=1,\ldots,n$. Let
$\varepsilon=(\varepsilon_1,\ldots,\varepsilon_n)^{\T}$.
$X=(x_{ij})_{n \times p} = (X_1,\ldots,X_p)$ is the $n\times p$ design
matrix and $Y=(y_1,\ldots,y_n)^{\T}$ is the response vector. To ensure
identifiability of $\{f_j$, $j=1,\ldots,p\}$, we assume
$E f_j(x_{ij})=0$.

For function $f_j$, let $\{\psi_{jk}, k=1,2, \ldots \}$ denote the
uniformly bounded basis functions with respect to the Lebesgue measure
on $[0,1]$.  Following \cite{ravikumar2009sparse}, we assume the
following smoothness condition.
\begin{condition}\label{C.1}
  For $j=1,\ldots, p$,
  $f_j\in \mathcal{S}_j$ where
  \begin{gather*}
    \mathcal{S}_j=\left\{f_j\in\mathcal{H}_j: f_j(x)=\sum_{k=1}^{\infty}\beta_{jk}\psi_{jk}(x),\sum_{k=1}^{\infty}\beta_{jk}^2k^4 \leq C^2\right\}
  \end{gather*}
  for some $0<C<\infty$, where $\mathcal{H}_j$ is a Hilbert space of
  mean zero square integrable functions with the inner product
  \(\langle f_j,f'_j\rangle=E f_j(x_{ij})f'_j(x_{ij})\), i.e.
  \(E f_j(x_{ij}) = 0\),
  \(\|f_j\|^2=\langle f_j, f_j \rangle <\infty\), and
  $\sup_x|\psi_{jk}(x)|\leq B$ for some $B$.  $\{\beta_{jk}$,
  $k=1, 2, \dots\}$ are the parameters corresponding to $f_j$.
\end{condition}

The standard form of the penalized additive model optimization problem is
\begin{align}
  \min_{f_1\in \mathcal{S}_1,\ldots, f_p\in \mathcal{S}_p} \sum_{i=1}^{n}\left\{y_i-\sum_{j=1}^{p}f_j(x_{ij})\right\}^2+J(f_{1},\ldots,f_{p}).
  \label{optmize1}
\end{align}
where $J$ is a sparsity-smoothness penalty.
In this paper, we restrict our discussion to the sparsity-inducing penalty
\begin{align*}
  J(f_{1},\ldots,f_{p})=\lambda_n\sum_{j=1}^{p}\sqrt{ \sum_{i=1}^n f_{j}^2(x_{ij})}.
\end{align*}

Following \cite{meier2009high}, we approximate $\{f_j$,
$j=1,\ldots, p\}$ by a cubic B-spline with a proper number of knots.
One possible choice would be to place $d_n-4$ interior knots at the
empirical quantile of $X_j$, i.e.,
\begin{align*}
  f_j(x)\approx{f}_{nj}(x)=\sum_{k=1}^{d_n}\beta_{jk}\psi_{jk}(x).
\end{align*}
With Condition \ref{C.1}, we can bound the truncation bias by
$\|f_j-f_{nj}\|^2=O(1/d_n^3)$. Let $h=\sum_{j=1}^pf_{j}$ and
$h_{n}=\sum_{j=1}^pf_{nj}$.  Let $S=\{j: f_j\neq 0\}$ be the active
set of variables and $s=|S|$ be its cardinality. It follows that
$\|h-h_{n}\|^2=O(s^2/d_n^3)$.

Let $\Psi_j$ denote the $n\times d_n$ B-spline basis matrix for $f_j$,
where $\Psi_j(i,k)=\psi_{jk}(x_{ij})$. Let $\beta_j$ denote the
corresponding coefficient vector $(\beta_{j1}, \ldots, \beta_{jd_n})$.
Then the optimization (\ref{optmize1}) can be reformulated as
\begin{align}
  \label{SpAM}
  \min_{\beta_1,\ldots,\beta_p}\left\|Y-\sum_{j=1}^{p}\Psi_j\beta_j\right\|^2+
  \lambda\sum_{j=1}^{p}\frac{1}{\sqrt{n}}\|\Psi_j\beta_j\|.
\end{align}

This group-wise variable selection problem can be solved by the standardized group lasso technique \citep{simon2012standardization}. The algorithm for standardized group lasso can be viewed as a special group lasso procedure after orthogonalization within each group, in which group lasso is computationally more intensive than lasso \citep{tibshirani1996regression}. Since its solution paths are not piecewise linear, the least angle regression (LARS) algorithm \citep{efron2004least} is not applicable. Instead, the block coordinate-wise descent-type algorithms \citep{hastie1990generalized, meier2008group, foygel2010exact, wood2011fast, yang2015fast} are common approaches.
Computational complexity is somewhat tricky to quantify since it largely depends on the number of iterations.
Since each spline block costs $O(nd_n)$ operations, $O(npd_n)$ calculations
are required for entire data in one pass. The number of back-fitting loops required for convergence is usually related to $p$.
As for the noniterative components, orthogonalization within each block can be solved by QR decomposition which costs $O(npd_n^2)$ operations.
Besides, compared with linear regression, memory footprint increases with the expanded spline basis functions $\Psi_j$ taking place of original $X_j$.
All these manifest that we need distributed learning to relieve stress from computation time and memory cost.

Before introducing our method, some additional notation is needed.
Let $\Psi=(\Psi_1,\ldots,\Psi_p)$ denote $n\times pd_n$ design matrix
of the B-spline bases and
$\beta=(\beta_1^{\T},\beta_2^{\T},\ldots,\beta_p^{\T})^{\T}$ be the
length-$pd_n$ coefficient vector.  If $A\subset\{1,\ldots,p\}$, we
denote the $n\times d_n|A|$ submatrix of $\Psi$ by $\Psi_A$ where for
each $j\in A$, $\Psi_j$ represents the submatrix in the corresponding
order.  Correspondingly, $\beta_A$ is the coefficients of $\Psi_A$.
For parallel computing, assume $X$ has been column-wisely partitioned
into $m$ groups, where \(m\) represents a pre-specified number of local machines one can access. If $X_j$ is assigned to the
$i$-th group and it is the $k$-th predictor in group $i$, we denote it
by $X_k^{(i)}$.  Note that there is a one-to-one mapping between the original predictor index $j$ and the $(i, k)$ pair. For convenience, define the mapping from the $(i, k)$ pair to the original index as $j = \zeta(i, k)$. 
We denote the $i$-th part of $X$ by
$X^{(i)}=(X^{(i)}_{1},X^{(i)}_{2},\ldots,X^{(i)}_{p_i})$ which are
stored in local machine $i$, $i=1,\ldots,m$ and its spline basis
matrix is denoted by $\Psi^{(i)}$.  Excluding $\Psi^{(i)}$, we denote
the remaining submatrix of $\Psi$ by $\Psi^{(-i)}$.  Let $S^{(i)}$
denote the true set of important variables in the $i$-th group, i.e.,
$S^{(i)}=\{\zeta(i, k): f_k^{(i)} \neq 0\}$, with $s_i=|S^{(i)}|$, and let
$S^{c(i)}=\{\zeta(i, k): f_k^{(i)} = 0\}$ denote its complement. Thus, $S$
is the union of $S^{(i)}$, $i=1,\ldots,m$, and $S^c$ denotes its
complement. $\Psi^{(i)}_S$ is the submatrix of $\Psi^{(i)}$ consisting
of spline basis of important predictors in the $i$-th group and
$\Psi^{(i)}_{S^c}$ is the basis matrix for the noise predictors.

\section{DDAC-SpAM Algorithm}
\label{sec3}
Since $X$ has already been column-wisely partitioned,
each local machine stores one subset of the predictors and $Y$.  Before the parallel variable selection,
let us begin with a decorrelation step for the additive model.

Reformulated as the linear combination of basis functions, we have
\begin{gather}
  Y=\Psi\beta +Z+\varepsilon.
  \label{equation}
\end{gather}
where $Z=(z_1,\ldots,z_n)^{\T}$ with $z_i=\sum_{j=1}^p[f_{j}(x_{ij})-f_{nj}(x_{ij})]$, $i=1,\ldots,n$.

The most intuitive way to reduce correlation is orthogonalizing the basis matrix $\Psi$ to make its columns uncorrelated by left-multiplication.
If $\Psi$  has full column rank with $n > p d_n$, we write $\Psi$ via singular value decomposition as $\Psi=UDV^{\T}$, where $U$ is a $n\times pd_n$ tall matrix with orthonormal columns, $D$ is a $pd_n\times pd_n$ diagonal matrix and $V$ is a $pd_n\times pd_n$ orthogonal matrix.
Then, we set $\widetilde{\Psi}=F\Psi$. Here, $F=UD^{-1}U^{\T}$.
It is easy to see that the columns of $\widetilde{\Psi}$ are
orthogonal. Actually, $F$ can be calculated in the central machine by
\begin{align*}
  \left(\sum_{i=1}^{m}\Psi^{(i)}\Psi^{(i)\T}\right)^{\frac{+}{2}},
\end{align*}
where the $n\times n$ matrix $\Psi^{(i)}\Psi^{(i)\T}$ is transmitted from local machine and $A^{+}$ denotes the Penrose-Moore pseudo-inverse of $A$.

Left multiplying $F$ on both sides of (\ref{equation}), we get
\begin{align*}
  FY=F\Psi\beta +FZ+F\varepsilon.
\end{align*}
It can be denoted as
\begin{align}
  \label{tildey}
  \widetilde{Y}=\widetilde{\Psi}\beta +\widetilde{Z}+\widetilde{\varepsilon}.
\end{align}
The spline basis matrix $\widetilde{\Psi}$ satisfies $\widetilde{\Psi}_{i}^{\T}\widetilde{\Psi}_{j}=0$ for any $i\neq j$ and $\widetilde{\Psi}_{i}^{\T}\widetilde{\Psi}_{i}=I_{d_n}$.

The group lasso working mechanism for the $i$-th data subgroup $\left\{\widetilde{Y}, \widetilde{\Psi}^{(i)}\right\}$ can be shown as follow.
Firstly, the optimization object (\ref{SpAM}) would be
\begin{gather}
  \label{subgroup}
  L(\beta^{(i)})=\left\|\widetilde{Y}-\sum_{k=1}^{p_i}\widetilde{\Psi}_k^{(i)}\beta_k^{(i)} \right\|^2+\lambda_n\sum_{k=1}^{p_i}\frac{1}{\sqrt{n}}\|\widetilde{\Psi}_k^{(i)}\beta_k^{(i)}\|.
\end{gather}

As shown in \cite{yuan2006model} and \cite{ravikumar2009sparse},  a solution to (\ref{subgroup}) satisfies
\begin{align*}
  \hat{\beta}_k^{(i)}=\left[1-\frac{\lambda_n}{\|P_k^{(i)}\|} \right]_{+}P_k^{(i)},
\end{align*}
where $P_k^{(i)}=\widetilde{\Psi}_{k}^{(i)\T}\widetilde{Y}$.

Combining with  (\ref{tildey}), we can derive that
\begin{align}
  P_k^{(i)}&=\widetilde{\Psi}_{k}^{(i)\T}
             \left(\widetilde{\Psi}^{(i)}_k \beta^{(i)}_k+\widetilde{\Psi}^{(i)}_{-k}\beta^{(i)}_{-k}+
             \widetilde{\Psi}^{(-i)}\beta^{(-i)}
             +\widetilde{Z}+\widetilde{\varepsilon} \right) \notag \\
           &=\beta^{(i)}_k+\widetilde{\Psi}_{k}^{(i)\T}\widetilde{\Psi}^{(i)}_{-k}\beta^{(i)}_{-k}+\widetilde{\Psi}_{k}^{(i)\T}\widetilde{\Psi}^{(-i)}\beta^{(-i)}
             +\widetilde{\Psi}_{k}^{(i)\T}\widetilde{Z}
             +\widetilde{\Psi}_{k}^{(i)\T}\widetilde{\varepsilon}\notag \\
           &=\beta^{(i)}_k+\widetilde{\Psi}_{k}^{(i)\T}\widetilde{Z}
             +\widetilde{\Psi}_{k}^{(i)\T}\widetilde{\varepsilon}
             \label{transform}
\end{align}
where $\widetilde{\Psi}^{(i)}_{-k}=(\widetilde{\Psi}^{(i)}_1,\ldots,\widetilde{\Psi}^{(i)}_{k-1},\widetilde{\Psi}^{(i)}_{k+1},\ldots,\widetilde{\Psi}^{(i)}_{p_i} )$
and
$\beta^{(i)}_{-k}=(\beta^{(i)\T}_1$,$\ldots$,$\beta^{(i)\T}_{k-1}$,$\beta^{(i)\T}_{k+1}$,$\ldots$, $\beta^{(i)\T}_{p_i} )^{\T}$.
Since the last two terms of (\ref{transform}) can be bounded by Condition \ref{C.1}, with a mild condition  for $\Psi\Psi^{\T}$ to be presented in Section~\ref{sec4}, $P_{k}^{(i)}$ converges to $\beta_k^{(i)}$ almost at the same rate as that with the full data.

When $pd_n\geq n$, SVD of $\Psi$ generates a $n\times n$ orthogonal matrix $U$, a $pd_n\times n$ matrix $V$ with only orthonormal columns and $D$ is a $n\times n$ diagonal matrix. Then $F$ becomes
$(\sum_{i=1}^{m}\Psi^{(i)}\Psi^{(i)\T})^{-\frac{1}{2}} $.
Although the columns of $\widetilde{\Psi}$ are not exactly mutually orthogonal, i.e., for some $i\neq j$,
$\widetilde{\Psi}_{i}^{\T}\widetilde{\Psi}_{j}\neq 0$, according to \cite{khatri1965some}, we have
$
E(\widetilde{\Psi}^{\T}\widetilde{\Psi})=(n/pd_n)I_{pd_n},
$
which means that any two columns of $\widetilde{\Psi}$ are orthogonal in expectation.
Thus, we can still apply the same decorrelation step to get a new response $\widetilde{Y}$ and the design matrix $\widetilde{\Psi}$.

\begin{figure}[htb!]
  \begin{center}
    \includegraphics[width=10cm]{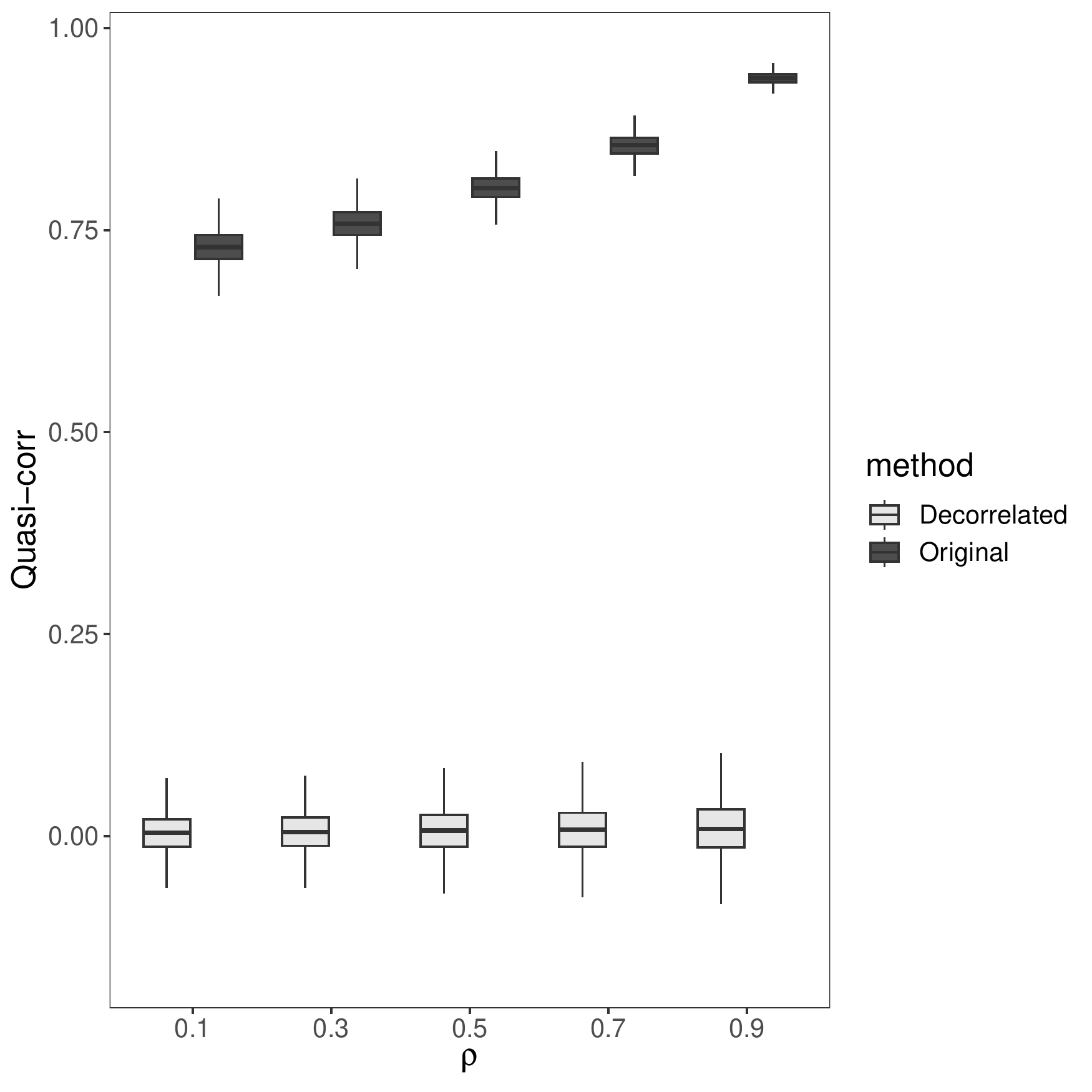}
  \end{center}
  \caption{Comparison of quasi-correlations for $\Psi$ and
    $\widetilde{\Psi}$. \label{Fig1}}
\end{figure}

The decorrelation operation mainly aims to lower the correlation
between the basis functions in different blocks. To show it visually,
we now present a simple example with $n< pd_n$.  In particular, we
sample $X$ from zero mean normal distribution with covariance matrix
$\Sigma = [\sigma_{ij}]$, where $\sigma_{ii}=1$, $\sigma_{ij}=\rho$
with $i\neq j$ and $(n,p)=(500,1000)$. A cubic B-spline with $d_n=5$
is used. We focus on comparing
$\tilde \rho_{ij}:= \tr(\widetilde{\Psi}^{\T}_i\widetilde{\Psi}_j)/
(\|\widetilde{\Psi}_i\|_F\|\widetilde{\Psi}_j\|_F)$ and
$\rho_{ij}: = \tr(\Psi^{\T}_i \Psi_j)/ (\|\Psi_i\|_F\|\Psi_j\|_F)$,
$1\leq i<j\leq p$. The difference between these two terms is affected
by the dependence between basis functions of different covariates. We
call them quasi-correlation here. Figure \ref{Fig1} shows the boxplots
of quasi-correlation before and after the decorrelation step when
$\rho$ increases. It can be seen that while $\rho_{ij}$ increases with
$\rho$, $\tilde \rho_{ij}$ is stable throughout the range of $\rho$ at
a very low level, which means the decorrelation step reduces
correlation between additive components in high-dimensional additive
model significantly.

After the decorrelation step, since
$\{\widetilde{\Psi}_k^{(i)}, k=1,\ldots,p_i\}$ is not exactly
column-orthogonal, we cannot apply the SpAM backfitting algorithm
\citep{ravikumar2009sparse} directly with
$\left\{\widetilde{Y}, \widetilde{\Psi}^{(i)}\right\}$ on the $i$-th
local machine. Following \cite{simon2012standardization}, we use a
method similar to the standardized group lasso to solve this problem.

Specifically,
we apply QR decomposition for each block $\widetilde{\Psi}^{(i)}_k$, $i=1,\ldots, m$, $k=1,\ldots, p_i$ into the product of an orthogonal matrix $\widetilde{Q}^{(i)}_k$ and an upper triangular matrix $\widetilde{R}^{(i)}_k$. Then,
the $i$-th local machine runs the SpAM backfitting algorithm with $\left\{\widetilde{Y}, \widetilde{Q}^{(i)}\right\}$ to
solve the following problem
\begin{align}
  \label{thetasolve}
  \hat{\theta}^{(i)}=\mathop{\arg\min}_{\theta^{(i)}=(\theta_1^{(i)\T},\ldots,\theta_{p_i}^{(i)\T})^{\T} }\frac{1}{n}\|\widetilde{Y}-\widetilde{Q}^{(i)}\theta^{(i)}\|^2+\lambda_n\sum_{k=1}^{p_i}\|\widetilde{Q}^{(i)}_k\theta^{(i)}_k\|, \quad i=1,\ldots,m
\end{align}
and select variables, where $\widetilde{Q}^{(i)}=(\widetilde{Q}^{(i)}_1,\ldots,\widetilde{Q}^{(i)}_{p_i})$.

The original coordinates $\hat{\beta}^{(i)}$ can be back-solved by
\begin{equation}
  \label{backsolve}
  \hat{\beta}^{(i)}_k=(\widetilde{R}_k^{(i)})^{-1}\hat{\theta}_k^{(i)}.
\end{equation}
However, this is unnecessary for the purpose of feature selection
since $\hat{\theta}^{(i)}_k=0$ implies $\hat{\beta}^{(i)}_k=0$. The
local machines only need to transfer the selected important variables
and their basis functions to the central machine.  The final estimates
of $\beta^{(i)}$ and $f_j(X_j)$ will be computed on the central
machine.

Let $\hat{S}^{(i)}=\{\zeta(i, k): \hat{\theta}_k^{(i)}\neq 0\}$  denote the $i$-th estimated set of important variables, for $i=1,\ldots,m$,
and $\hat{S}$ be their union.
The details of DDAC-SpAM are provided in Algorithm \ref{algo}.

\begin{algorithm}[t]
  \caption{Divide, decorrelate and conquer SpAM (DDAC-SpAM)}	
  \label{algo}
  \KwIn{$Y$, $X$, $d_n$, the number of machines $m$,  the ridge regularization parameter $r$.}
  On the central machine,  store and standardize $Y$ to get $\underline{Y}$\;
  Randomly divide predictors into $m$ parts: $X^{(1)}, \ldots, X^{(m)}$ and allocate $(\underline{Y},X^{(i)})$ to the $i$-th local machine for $i=1,\ldots, m$\;
  On the $i$-th local machine, $i=1,\dots,m$, generate spline basis matrix $\Psi^{(i)}$ for $X^{(i)}$,   standardize every column of $\Psi^{(i)}$  to get $\underline{\Psi}^{(i)}$ and  transmit $\underline{\Psi}^{(i)}\underline{\Psi}^{(i)\T}$ to the central machine\;
  On the central machine, compute $F=(\sum_{i=1}^{m}\underline{\Psi}^{(i)}\underline{\Psi}^{(i)\T}+rI_n)^{-1/2}$ and transmit it to the local machines\;
  On the $i$-th local machine, $i=1,\dots,m$, compute $\widetilde{\Psi}^{(i)}=F\underline{\Psi}^{(i)}$ and $\widetilde{Y}=F\underline{Y}$\;
  On the $i$-th local machine, $i=1,\dots,m$, (a) perform the QR factorization $\widetilde{\Psi}^{(i)}_k=\widetilde{Q}_k^{(i)}\widetilde{R}_k^{(i)}$, for $k=1, \ldots, p_i$; (b) run the SpAM backfitting algorithm to solve (\ref{thetasolve}); (c) push $\hat{S}^{(i)}$  and $\Psi_{\hat{S}^{(i)}}$ to the central machine\;
  On the central machine, combine $\Psi_{\hat{S}^{(i)}}$, $i=1,\ldots,m$, to get $\Psi_{\hat{S}}$. Apply ridge regression on $(Y, \Psi_{\hat{S}})$  and get $\hat{\beta}_{\hat{S}}$\;
  \KwOut{$\hat{S}$  and $\hat{f}_j$, $j\in \hat{S}$.}
\end{algorithm}

\begin{remark}
  Steps 1 and 2 are used for data initialization and division. Steps 3--5 are the decorrelation steps. Step 6 and 7 are distributed feature selection and final refinement steps, respectively.
  In Step 4, we use $\sum_{i=1}^{m}\underline{\Psi}^{(i)}\underline{\Psi}^{(i)\T}+rI_n$ instead of $\sum_{i=1}^{m}\underline{\Psi}^{(i)}\underline{\Psi}^{(i)\T}$ for robustness.
  Besides, using ridge regression in Step 7 instead of ordinary least squares is also for robustness.
\end{remark}

Now, we analyze the computational complexity and memory consumption of
DDAC-SpAM. For convenience, we assume that the $p$ features are evenly
distributed to $m$ parts. Excluding spline interpolation, the costs of
the decorrelation operation and QR factorization are
$O(n^3+n^2pd_n /m +n^2 m )$ and $O(npd_n^2/m)$ per local machine,
respectively, so the total cost is $O(n^3+n^2pd_n /m +n^2 m )$. For parallel
estimation, within each iteration of the SpAM backfitting algorithm,
$O(npd_n/m)$ calculations are required. Assume the number of loops is
$k$.  So the total computational cost is
$O(n^3+n^2pd_n /m +n^2 m + knpd_n/m)$ for DDAC-SpAM, compared with
$O(knpd_n+npd_n^2)$ for SpAM on a single machine.  Meanwhile, the
memory consumption of every local machine through the entire algorithm
is decreased roughly by a factor of $m$.  As shown above, DDAC-SpAM
can significantly speed up computation and relax memory
requirements. This will be demonstrated in the numerical study
section.

\section{Statistical Inference via Debiased DDAC-SpAM}
\label{sec:stat-infer-via}

In the previous section, we have developed the DDAC-SpAM algorithm for
feature selection of the sparse additive model. With the selected
variables, we further apply ridge regression to obtain an estimation
of the coefficients \(\beta\) in (\refeq{equation}). In this section,
we study the statistical inference for \(\beta\).

Naturally associated with the sparse additive model
\(y_i=\sumjp f_j(x_{ij})+\varepsilon_i\), we focus on the fundamental
hypothesis testing problem \(H_0\): \(f_j = 0\) versus \(H_1\):
\(f_j \neq 0\) for some \(1 \leq j \leq p\). In terms of the B-spline
basis expansion under our distributed computing setting, this is
equivalent to
\begin{equation}
  \label{test}
  H_0: \beta^{(i)}_k = 0 \textrm{ vs. } H_1: \beta^{(i)}_k \neq 0
\end{equation}
for \(1 \leq i \leq m\) and \(1 \leq k \leq p_i\).

Based on recent developments in high-dimensional inference for linear
models \citep{cai2022sparse,van2014asymptotically}, we construct the
following debiased DDAC-SpAM estimator
\begin{equation}
  \label{4.1}
  \hat\beta^{u} = \hat\beta + \frac{pd_n}{n}
  \widetilde{\Psi}^{\T}\left(\widetilde Y -
    \widetilde\Psi\hat\beta\right), 
\end{equation}
where \(\hat \beta\) is the DDAC-SpAM estimate of \(\beta\) obtained
from (\refeq{backsolve}). Restricting ourselves to the \(k\)-th
variable in the \(i\)-th machine, \(\hat\beta_k^{u(i)}\) enjoys the
property that its scaled and decorrelated version 
$\widehat{M}^{(i)}_k \left(\hat\beta_k^{u(i)} - \beta^{(i)}_k\right)$
approximately follows a \(d_n\)-dimensional standard normal
distribution, where
\begin{equation}
  \label{4.2}
  \widehat{M}^{(i)}_k = (\tildePsijT F F^{\T} \tildePsij)^{-1/2} n/(pd_n \hat\sigma),
\end{equation}
in which $\hat \sigma$ is an estimate for $\sigma$
(Theorem~\ref{thm2}). Therefore, we define our test statistic
\(\mathcal{T}^{(i)}_k: = \|\widehat{M}^{(i)}_k
\hat\beta_k^{u(i)}\|^2\), which follows \(\chi^2(d_n)\) asymptotically
under \(H_0\). The notation $\mathcal{T}_j := \mathcal{T}^{(i)}_k$ is
also used when $j=\zeta(i,k)$. With significance level \(\alpha_0\),
we reject the null hypothesis of \eqref{test} when
\(\mathcal{T}^{(i)}_k > F^{-1}_{d_n}(1 - \alpha_0)\), where
\( F_{d_n}\) is the cumulative distribution function for
\(\chi^2(d_n)\). The detailed testing procedure is presented in
Algorithm~{\ref{algo2}}.

\begin{remark}
  In \cite{cai2022sparse} and \cite{van2014asymptotically}, the
  debiased estimator is constructed as
  \(\hat\beta^{u} = \hat\beta + \widehat \Theta \Psi^{\T}( Y -
  \Psi\hat\beta)/n\), where \(\widehat \Theta\) is an estimate of
  \((\Psi^{\T}\Psi/n)^{-1}\).  When working with decorrelated
  \(\widetilde \Psi\) and \(\widetilde Y\), we have
  \((\widetilde{\Psi}^{\T}\widetilde{\Psi}/n)^{-1} \approx pd_n
  I_{pd_n}\) \citep{khatri1965some}, which leads to the definition in
  (\refeq{4.1}).
\end{remark}

\begin{algorithm}[htb]
  \caption{Inference via Debiased DDAC-SpAM (\(H_0: \beta^{(i_0)}_{k_0} = 0 \textrm{ vs. } H_1: \beta^{(i_0)}_{k_0} \neq 0\))}
  \label{algo2}
  \KwIn{$Y$, $X$, $d_n$, the number of machines $m$,  the ridge regularization parameter $r$, the significance level \(\alpha_0\).}
  On the central machine,  store and center $Y$ to get $\underline{Y}$\;
  Steps 2-5 of Algorithm~\ref{algo}\;
  On the $i$-th local machine, $i=1,\dots,m$, perform the QR factorization $\widetilde{\Psi}^{(i)}_k=\widetilde{Q}_k^{(i)}\widetilde{R}_k^{(i)}$, for $k=1, \ldots, p_i$, run the SpAM backfitting algorithm to solve (\ref{thetasolve}), compute \(\hat{\beta}^{(i)}\) by \(\hat{\beta}^{(i)}_k=(\widetilde{R}_k^{(i)})^{-1}\hat{\theta}_k^{(i)}\), and push \(\widehat{Y}^{(i)} = \underline{\Psi}^{(i)} \hat{\beta}^{(i)}\) to the central machine\;
  On the central machine, fetch $\hat{\beta}^{(i_0)}_{k_0}$ and \(\widetilde{\Psi}^{(i_0)}_{k_0}\) from the \(i_0\)-th machine, compute \(\hat \e = Y - \sum_{i = 1}^m \widehat{Y}^{(i)}\), \(\hat \sigma = \|\hat \e\|/\sqrt{n}\), \(\widehat M^{(i_0)}_{k_0} = (\widetilde{\Psi}^{(i_0)\T}_{k_0} F F^{\T} \widetilde{\Psi}^{(i_0)}_{k_0})^{-1/2} n/(pd_n \hat \sigma)\), \(\hat\beta^{u(i_0)}_{k_0} = \hat\beta^{(i_0)}_{k_0} + pd_n\widetilde{\Psi}^{(i_0)\T}_{k_0}F \hat \e/n\) and \(\mathcal{T}^{(i_0)}_{k_0} = \|\widehat M^{(i_0)}_{k_0} \hat\beta_{k_0}^{u(i_0)}\|^2\)\;
  \KwOut{``Reject'' if \(\mathcal{T}^{(i_0)}_{k_0} > F^{-1}_{d_n}(1 - \alpha_0)\); ``Accept'' otherwise.}
\end{algorithm}

\section{Theoretical Results}
\label{sec4}
In this section, we provide the theoretical framework for DDAC-SpAM to
show it is variable selection consistent (\textit{sparsistent}) under
mild conditions. On top of that, we further derive the asymptotic
distribution for the debiased DDAC-SpAM estimator.

For results in this section, we will treat $X$ as random. When $p d_n\geq n$, let the spline interpolation $\Psi=UDV^{\T}$, where $U$ is a $n\times n$ orthogonal, $D$ is a $n\times n$ diagonal matrix and $V$ satisfies $V^{\T}V=I_n$. $\widetilde{\Psi}=F\Psi=UV^{\T}$ satisfies
$\widetilde{\Psi}\widetilde{\Psi}^{\T}=I_n$. All $n\times pd_n$ matrices whose rows are orthonormal (e.g. $\widetilde{\Psi}$) form \emph{Stiefel manifold} $\mathcal{V}(n,pd_n)$ \citep{downs1972orientation}.

For clarity, we review the definition of uniform distribution for a random matrix.
\begin{definition}\citep{chikuse2003statistics}
  A random $n\times p$ matrix $H$ is uniformly distributed on
  $\mathcal{V}(n,p)$, written $H\sim \mathrm{Uniform}$
  $ (\mathcal{V}(n,p))$, if $H$ has the same distribution as $HO$ for
  any fixed $p\times p$ orthogonal matrix $O$.
\end{definition}

Besides Condition \ref{C.1}, we further make the following assumptions for $\Psi$.

\begin{condition}
  $\widetilde \Psi\sim \mathrm{Uniform}$ $(\mathcal{V}(n,pd_n))$.
  \label{C.2}
\end{condition}

We allow the minimum eigenvalue of $\Psi\Psi^{\T}$ decay with sample
size at a certain rate.
\begin{condition}
  $\rP(\lambda_{\min}(\Psi\Psi^{\T}/pd_n)>\delta n^{\alpha-1})
  \geq 1 - \exp (-\xi n^{\gamma}), \ \textrm{for some} \ 0<\alpha
  \leq 1$ and \(\delta, \xi, \gamma > 0\).
  \label{C.3}
\end{condition}

Similar conditions to Condition \ref{C.2} have been imposed on the
design matrix of linear model \citep{jia2015preconditioning}.
Condition \ref{C.3} is related to Theorem 2 in
\cite{ravikumar2009sparse} in which they require the eigenvalues of
$n^{-1}\Psi^{\T}\Psi$ to be bounded by constants. When the number of
covariates diverges with $n$, Condition \ref{C.3} is more general than
theirs.

We also assume that the truncation size $d_n$, regularization
parameter $\lambda_n$ and the number of important variables $s$
satisfy

\begin{condition}
  \(d_n \to \infty\), \(\tilde \lambda_n \to 0\),
  \(\tilde \lambda_n^{-2} n^{-1}s^2d_n \to 0\),
  $\tilde \lambda_n^{-2}n^{1-\alpha}s^2d_n^{-3}\rightarrow 0$,
  \(\tilde \lambda_n^{-2}n^{-\alpha}sd_n \to 0\) and
  $\sqrt{s}\tilde\lambda_n/\rho_n \to 0$, where
  \(\tilde \lambda_n = \sqrt{pd_n} \lambda_n\) and
  $\rho_n=\min_{j\in S}\|\beta_j\|_\infty$.
  \label{C.4}
\end{condition}

Similar conditions were assumed in many high-dimensional additive
model variable selection literatures, such as condition (B2) for
Theorem 3 in \cite{huang2010variable}.

Additionally,  assume
\begin{condition}
  $p=o(\exp(sd_n))$.
  \label{C.5}
\end{condition}

To clarify the implications of Conditions \ref{C.4} and \ref{C.5},
assume the number of important variables is bounded, i.e., $s=O(1)$.
Then, in practice, we can set $d_n\asymp n^{1/5}$. The order of
dimension $p_n$ can be as large as $o(\exp(n^{1/5}))$.  If
$1/\rho_n=o(n^{\alpha/2-1/5}/\log{n})$, a suitable choice for the
regularization parameter $\tilde\lambda_n$ would be
$n^{1/5-\alpha/2}\log{n}$ for some $2/5<\alpha\leq 1$.

If we impose a slightly more strict requirement on \(\alpha\) (e.g.,
\(3/5 < \alpha \leq 1\)), then the sparsity \(s\) is allowed to
increase with \(n\). For example, assuming \(s \asymp n^{1/10}\),
\(p = o(\exp(n^{3/10}))\), \(d_n\asymp n^{1/5}\) and
\(1/\rho_n=o(n^{\alpha/2-7/20}/\log n)\), we can choose the
regularization parameter
\(\tilde \lambda_n \asymp n^{3/10 - \alpha/2}\log n\) so that both
Conditions 4 and 5 are satisfied.

The key of our algorithm is to reduce the correlation between predictors which leads to a milder constraint for the correlation structure of variables or $f_j(X_j)$
within Theorem 1 than before.
It can be reflected in two aspects. First, there is no assumption for the correlation between important variables that are distributed to different local machines, since the bound of this kind of correlation is reduced to
\begin{gather*}
  \rP\left\{\left\|\frac{pd_n\tildePsiT\tildePsino}{n}\right\|\leq \frac{C_1\tilde\lambda_n}{\sqrt{s}}\right\}\rightarrow 1,
\end{gather*}
for some $C_1>0$, as shown in Lemma S.2 in the supplementary material
of this paper.  Second, there is no assumption for the correlation
between important variables and irrelevant variables.  Specifically,
we do not need a version of \textsl{irrepresentable condition}
\citep{zhao2006model} for selection consistency of DDAC-SpAM. In previous
works, such as \cite{ravikumar2009sparse}, this kind of condition can
be formulated as an upper bound for
$$\max_{j\in S^c}\left\|\left(\frac{1}{n}\Psi_j^{\T}\Psi_S\right)\left(\frac{1}{n}\Psi_S^{\T}\Psi_S\right)^{-1}\right\|, $$
while this bound exists in our work by
\begin{gather*}
  \rP\left\{\max_{j\in S^{c(i)}}\left\|\frac{pd_n\tildePsiT\tildePsij}{n}\right\|\leq\frac{C_2}{\sqrt{s}},\left\|\left(\frac{pd_n}{n}\tildePsiT\tildePsi\right)^{-1}\right\|\leq C_3 \right\}\rightarrow 1,
\end{gather*}
for some $C_2,C_3>0$.  The details and proof of these results can be
found in Lemmas S.2, S.3, S.4 of the supplementary material.

\begin{theorem}
  Assuming Conditions \ref{C.1}-\ref{C.5} hold and \(s_i > 0\), the
  following inequality holds for sufficiently large \(n\):
  $$\rP\left(\hat{S}^{(i)}=S^{(i)}\right)\geq 1-\exp(-\xi n^\gamma) - 16(p_i - s_i + 1)\exp(-sd_n) \rightarrow 1, $$
  i.e., the local estimator on machine $i$ is sparsistent.
\end{theorem}

The convergence rate of \(\rP(\hat{S}^{(i)}=S^{(i)})\) is controlled by two
\(o(1)\) terms, where \(\exp(-\xi n^\gamma)\) directly follows from
Condition \ref{C.3}, and \(16(p_i - s_i + 1)\exp(-sd_n)\), as a
combined rate, has connections to both the Gaussian assumption of
\(\e\) and the uniform assumption in Condition \ref{C.2}. The
additional technical condition \(s_i > 0\) is not critical in our numerical studies,
as we will show in the following sections. The proof of
Theorem 1 is provided in the supplementary material.

After aggregating the results from local machines, we can derive the
following corollary.
\begin{corollary}
  Under Conditions \ref{C.1}-\ref{C.5}, the central estimator is sparsistent:
  $$\rP\left(\hat{S}=S\right)\geq 1-\exp(-\xi n^\gamma) - 16(p - s + 1)\exp(-sd_n) \rightarrow 1. $$
\end{corollary}

Next, we will establish the theoretical foundation of
Algorithm~\ref{algo2}. To ensure our test statistic \(\mathcal{T}\)
follow the desired chi-squared distribution asymptotically, we need
two additional conditions.
\begin{condition}
  $\rP(\lambda_{\max}(\Psi\Psi^{\T}/pd_n) \leq \delta') \geq 1 - \exp
  (-\xi' n^{\gamma'})$ for some \(\delta', \xi', \gamma' > 0\).
  \label{C.6}
\end{condition}

\begin{condition}
  \(ns^2d_n^{-3} \to 0\) and
  \(\sqrt{s} \tilde \lambda_n n^{1/4} \to 0\).
  \label{C.7}
\end{condition}

Condition~\ref{C.6} resembles Condition~\ref{C.3} and requires the
eigenvalues of \(\Psi \Psi^{\T}\) to be bounded from
above with high probability. Condition~\ref{C.7} strengthens Condition~\ref{C.4} to make the
more challenging statistical inference a feasible task. One immediate
implication of Condition~\ref{C.7} is that more knots are needed for
the B-spline basis. With \(s = O(1)\), it is necessary that
\(d_n \gg n^{1/3}\). For instance, we may set
\(d_n \asymp n^{1/3}\log n\), which along with
\(\tilde\lambda_n \asymp n^{1/6 - \alpha/2} \log n\),
\(1/\rho_n=o(n^{\alpha/2 - 1/6}/\log{n})\) and
\(p = o(\exp(n^{1/3}\log n))\) would satisfy all requirements of
Conditions~\ref{C.4}, \ref{C.5}, and \ref{C.7} for
\(5/6 < \alpha \leq 1\).

\begin{theorem}
\label{thm2}
  Assuming Conditions \ref{C.1}-\ref{C.7} hold and \(s_i > 0\), the
  debiased DDAC-SpAM estimator \(\hat\beta^{u}\) has the following
  asymptotic distribution:
  \[
    \widehat{M}_k^{(i)} \left(\hat\beta^{u(i)}_k - \beta^{(i)}_k\right)
    \xrightarrow[]{d} \mathcal{N}(0, I_{d_n}),
  \]
  for all \(1 \leq i \leq m\) and \(1 \leq k \leq p_i\), where
  \(\widehat{M}_k^{(i)} = (\tildePsijT F F^{\T} \tildePsij)^{-1/2}
  n/(pd_n \sigma)\) is \eqref{4.2} with $\hat\sigma$ replaced by its
  true value $\sigma$. In addition, the test statistic
  \[
    \mathcal{T}_k^{(i)} = \|\widehat{M}_k^{(i)} \hat\beta_k^{u(i)}\|^2 \xrightarrow[]{d}
    \chi^2(d_n)
  \]
  when \(\beta^{(i)}_k = 0\).
\end{theorem}

\section{Simulation Studies}
\label{sec5}
\subsection{Performance Comparison Under Different Correlation Structures and Distributions}
\label{perf_comp}
To study the performance of the DDAC-SpAM procedure on simulated data
sets, we divide feature space evenly and randomly. Since the results are stable for different choices of $r$, we fix it to be 1 throughout the numerical studies for simplicity. For each $f_j$, we use a cubic B-spline
parameterization with $d_n=\lceil 0.1n^{1/3} \log n\rceil$ according
to the discussions after Condition~\ref{C.7}.  For comparison, we
include the full data SpAM with a ridge refinement (SpAM) and SpAM
with separated feature space without decorrelation (DAC-SpAM). The
oracle method that employs Step 7 of Algorithm~\ref{algo} with
\(\Psi_{\hat S}\) replaced by \(\Psi_S\) is also included as the
benchmark. We report the false positives (FP), false negatives (FN),
the mean squared error $\|\hat{h}-h\|^2 $ (MSE) and computational time
(Time). We use \verb"gglasso" \citep{YangZougglasso} and \verb"glmnet"
\citep{FriedmanHastieglmnet} with five-fold cross-validation to fit
group lasso and ridge regression, respectively. We consider an exponentially-decay sequence for $\lambda_n$, whose value varies from $\lambda_n^{(1)} > \lambda_n^{(2)} > \ldots > \lambda_n^{(500)}$, where $\lambda_n^{(1)}$ is the smallest $\lambda$ value such that  all coefficient estimates are zero  and $\lambda_n^{(500)} = 0.001 \lambda_n^{(1)}$.

We define the signal-to-noise ratio
$$\textrm{SNR}=\frac{\textrm{var}(h(X))}{\textrm{var}(\varepsilon)}. $$

\textsl{\textbf{Independent Predictors.}} We first consider the case with independent predictors. Two examples where predictors follow a uniform distribution and normal distribution are analyzed,  respectively.

\textsc{Example 1} ($SNR\approx15$). Following Example 1 in \cite{meier2009high},
we generate the data from the following additive model:
$$y_i=2g_1(x_{i1})+1.6g_2(x_{i2})-4g_3(x_{i3},2)+g_4(x_{i4})+1.5\varepsilon_i, $$
with
\begin{equation*}
  g_1(x)=x, \quad  g_2(x)=x^2-\frac{25}{12},\quad  g_3(x,\omega)=\sin(\omega x), \quad
  g_4(x)=e^{-x}-2/5\cdot\sinh(5/2),
\end{equation*}
where the covariates $X=(X_1,\ldots, X_p)$ are simulated from independent Uniform (-2.5, 2.5)  and $\varepsilon_i \stackrel{i.i.d.}{\sim} \mathcal{N}(0,1)$.

\textsc{Example 2} ($SNR\approx15$). In this example, the covariates $X=(X_1,\ldots, X_p)$ are simulated from independent standard normal distribution. The model is
$$y_i=5g_1(x_{i1})+2.1g_5(x_{i2})+13.2g_6(x_{i3},\frac{\pi}{4})+17.2g_7(x_{i4},\frac{\pi}{4})+2.56\varepsilon_i, $$
with
\begin{gather*}
  g_5(x)=(x-1)^2,\quad
  g_6(x,\omega)=\frac{\sin(\omega x)}{2-\sin(\omega x)},\\
  g_7(x,\omega)=0.1\sin(\omega x)+0.2\cos(\omega x)+0.3\sin^2(\omega x)+0.4\cos^3(\omega x)+0.5\sin^3(\omega x),
\end{gather*}
  and $\varepsilon_i \stackrel{i.i.d.}{\sim} \mathcal{N}(0,1)$.

\textsl{\textbf{Dependent Predictors.}} For dependent predictors with different distributions, we investigate three different correlation structures.

\textsc{Example 3} ($SNR\approx6.7$). Following Example 3 in \cite{meier2009high}, the covariates are generated with the following random-effects model:
\begin{gather*}
  X_j=\frac{W_j+tU_{\lceil j/20\rceil}}{1+t}, j=1,\ldots,p,
\end{gather*}
where $W_1,\ldots, W_p, U_1,\ldots,U_{\lceil p/20\rceil} \stackrel{i.i.d.}{\sim}$ Uniform $(0, 1)$.  By construction, the $p$ predictors are partitioned into segments of size 20. Variables in different segments are independent while the variables in each segment are dependent through the shared $U$ variable. As a result, the correlation strength within each segment is controlled by $t$. Here, we set $t=1.5$, leading to a correlation between $X_i$ and $X_j$ to be 0.6 when they are in the same segment.
The model is
$$y_i=2.5g_1(x_{i1})+2.6g_5(x_{i2})+g_6(x_{i3},2\pi)+g_7(x_{i4},2\pi)+0.3\varepsilon_i, $$
 where $\varepsilon_i \stackrel{i.i.d.}{\sim} \mathcal{N}(0,1)$.

\textsc{Example 4} ($SNR\approx6.7$). The setting is the same as Example 3 except that $X_j=(W_j+tU)/(1+t)$ and
$U\sim $ Uniform (0,1). We set $t=1.5$ leading to the pairwise correlation of all covariates being 0.6.

\textsc{Example 5} ($SNR\approx18$).  The covariates are generated
according to a multivariate normal distribution with zero mean and
covariance matrix $\Sigma=[\sigma_{ij}]$, where \(\sigma_{ii} = 1\)
and \(\sigma_{ij} = 0.5\) for \(i \neq j\). $Y$ is generated with
$$y_i=2.5g_1(x_{i1})+g_5(x_{i2})+6.5g_6(x_{i3},\frac{\pi}{4})+8.5g_7(x_{i4},\frac{\pi}{4})+1.2\varepsilon_i, $$
 where $\varepsilon_i \stackrel{i.i.d.}{\sim} \mathcal{N}(0,1)$.

For all examples, the feature dimension and the sample size are fixed at $p=10,000$ and $n=500$ respectively, which leads to $d_n = 5$. The number of machines is fixed as $m=20$. We run each simulation for 100 times and report the average performance in Table \ref{table1}. 

Several conclusions can be drawn from Table \ref{table1}. In Examples 1 and 2 where all variables are independent, DDAC-SpAM and DAC-SpAM perform the best, closely mimicking the performance of Oracle in terms of MSE. The possible reason for the similar performance between DDAC-SpAM and DAC-SpAM is that the decorrelation step is not necessary under this independent setting. On the other hand, SpAM tends to select more irrelevant variables. This shows that for independent variables,
distributed feature selection can enhance the selection accuracy. Also, we can see that both DDAC-SpAM and DAC-SpAM take much less time than SpAM, showing the power of distributed computing. 
In Examples 3-5 where the variables are dependent, the performances of both SpAM and DAC-SpAM
deteriorate, possibly due to the violation of the irrepresentable condition.
On the contrary, DDAC-SpAM is far less affected and achieves the overall best performance. In particular, it has much fewer false positives than the other two methods. This shows that the decorrelation step can handle such kinds of strong correlation structures. Also it leads to a smaller MSE than SpAM and DAC-SpAM for Examples 4 and 5. This shows the importance of the decorrelation step when there exists correlation among features. Similar to Examples 1 and 2, we observe DDAC-SpAM and DAC-SpAM are much faster to compute than SpAM. Interestingly, DDAC-SpAM takes significantly less time to compute than DAC-SpAM, possibly due to a faster SpAM fitting after decorrelation. 

\begin{table}[!htbp]
  \caption{Average false positive (FP), false negative (FN), mean squared error (MSE),  time (in seconds) over 100 repetitions and their standard deviations (in parentheses).  \label{table1}}
  \begin{center}
    \begin{tabular}{lccccc}
      Model & Method & \multicolumn{1}{c}{FP} & \multicolumn{1}{c}{FN} & \multicolumn{1}{c}{MSE} & \multicolumn{1}{c}{Time} \\\hline
        Example 1 & DDAC-SpAM & 0.02 (0.14) & 0.00 (0.00) & 0.465 (0.09) & 42.96 (2.55) \\ 
        & DAC-SpAM & 0.04 (0.20) & 0.00 (0.00) & 0.466 (0.09) & 42.05 (2.97) \\ 
        & SpAM & 1.22 (3.55) & 0.00 (0.00) & 0.562 (0.20) & 1524.47 (348.38) \\ 
        & Oracle & NA (NA) & NA (NA) & 0.464 (0.09) & NA (NA) \\ 
        Example 2 & DDAC-SpAM & 0.03 (0.22) & 0.00 (0.00) & 2.369 (0.59) & 42.60 (2.08) \\ 
        & DAC-SpAM & 0.01 (0.10) & 0.00 (0.00) & 2.365 (0.59) & 41.70 (1.95) \\ 
        & SpAM & 0.61 (1.47) & 0.00 (0.00) & 2.496 (0.64) & 1521.43 (243.08) \\ 
        & Oracle & NA (NA) & NA (NA) & 2.367 (0.59) & NA (NA) \\ 
        Example 3 & DDAC-SpAM & 3.59 (3.58) & 0.09 (0.29) & 0.036 (0.03) & 42.02 (2.18) \\ 
        & DAC-SpAM & 4.67 (3.15) & 0.07 (0.26) & 0.035 (0.02) & 41.48 (2.20) \\ 
        & SpAM & 6.58 (6.39) & 0.00 (0.00) & 0.036 (0.01) & 1542.65 (302.87) \\ 
        & Oracle & NA (NA) & NA (NA) & 0.026 (0.00) & NA (NA) \\ 
        Example 4 & DDAC-SpAM & 0.01 (0.10) & 0.03 (0.17) & 0.030 (0.02) & 43.44 (2.39) \\ 
        & DAC-SpAM & 41.23 (14.21) & 0.00 (0.00) & 0.055 (0.01) & 76.73 (8.12) \\ 
        & SpAM & 22.23 (11.67) & 0.00 (0.00) & 0.047 (0.01) & 1386.66 (243.80) \\ 
        & Oracle & NA (NA) & NA (NA) & 0.026 (0.00) & NA (NA) \\ 
        Example 5 & DDAC-SpAM & 0.20 (1.15) & 0.01 (0.10) & 0.680 (0.39) & 42.07 (2.51) \\ 
        & DAC-SpAM & 18.81 (13.69) & 0.00 (0.00) & 0.887 (0.16) & 90.10 (6.64) \\ 
        & SpAM & 11.92 (9.00) & 0.00 (0.00) & 0.832 (0.15) & 1514.38 (396.94) \\ 
        & Oracle & NA (NA) & NA (NA) & 0.643 (0.13) & NA (NA) \\ 
    \end{tabular}
  \end{center}
\end{table}

\subsection{Performance Comparison with Varying Number of Machines}

Corollary 1 indicates the sparsistency property of DDAC-SpAM is
irrelevant to the number of machines \(m\). This is because in the
ideal scenario where the decorrelation step produces perfectly
independent covariates, the aggregated result from the local machines
is identical to the full data estimator. In reality, correlation
between variables can never be fully eliminated, in which case 
a large \(m\) value mainly has two effects. First, it increases 
bias by distributing correlated variables into different machines,
which can potentially hurt the performance of DDAC-SpAM\@. Second, it
tends to separate correlated important variables, encouraging their
simultaneous selection.

In this experiment, we analyze the impact of the number of machines on the performance of DDAC-SpAM with simulated data. We fix the sample size
$n = 500$ and the dimension $p = 10,000$, and vary the number of
machines $m$ from 1 to 200 ($m=1,10,20,100,200$). Naturally, as $m$
increases, each machine has a lower local dimension. The data is
generated using Example 4 in Section~\ref{perf_comp}.

\begin{figure}[t]
  \begin{center}
    \includegraphics[width=12cm]{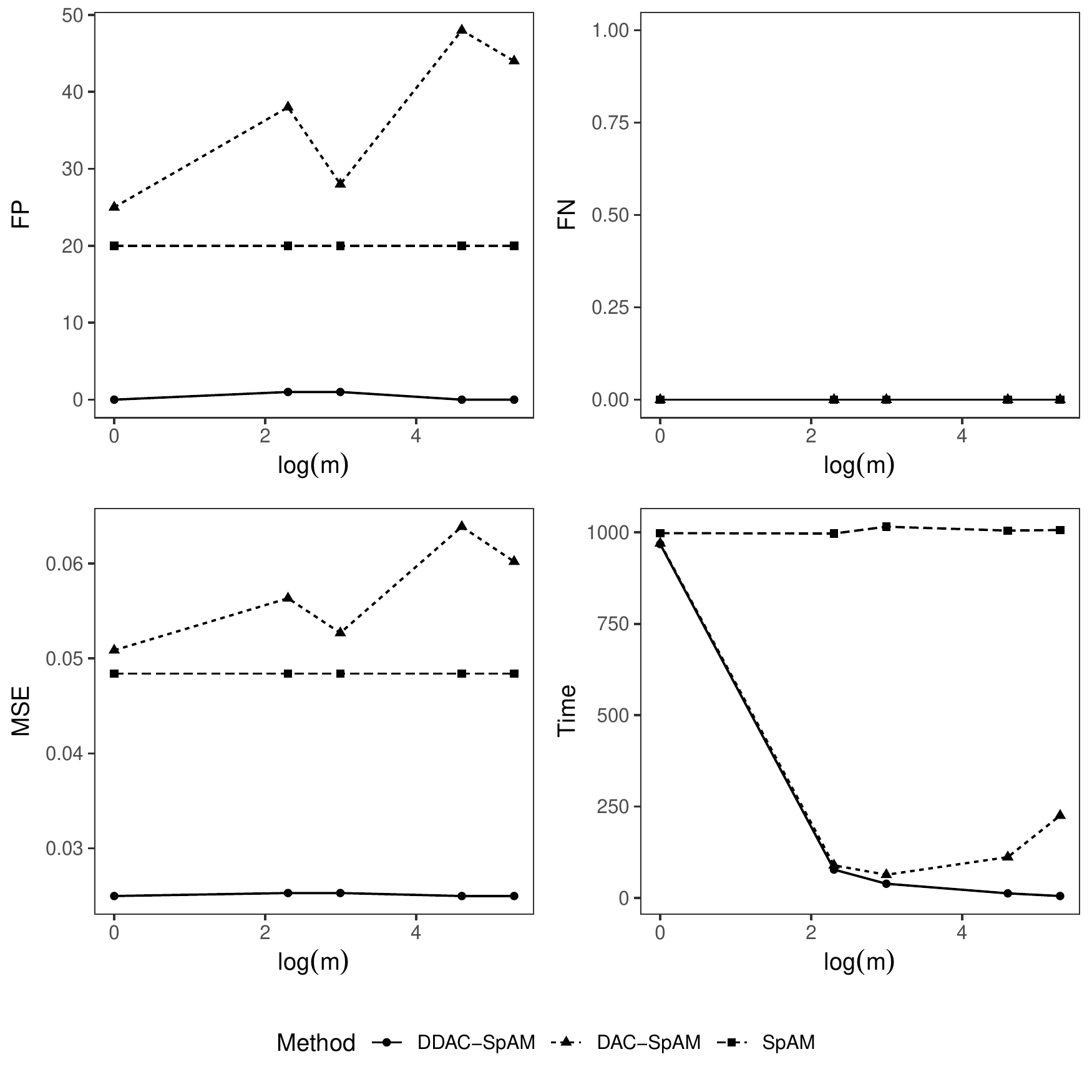}
  \end{center}
  \caption{Performance of DDAC-SpAM, DAC-SpAM and SpAM with different number of local machines. \label{Fig2}}
\end{figure}

We summarize the results in Figure~\ref{Fig2}. First, we observe that
all three methods capture nearly all important variables. Compared
with DAC-SpAM and SpAM, DDAC-SpAM has the smallest number of false
positive variables and lowest estimation error. While DAC-SpAM suffers
from high correlation between features when the computation is
distributed, i.e., \(m > 1\), the feature selection and prediction
performance of DDAC-SpAM is stable throughout the range of
\(m\). Besides, thanks to the distributed framework, the time
consumption of DDAC-SpAM and DAC-SpAM decreases as $m$ increases.
Although decorrelation increases the computational complexity which is
evident when the data set is not partitioned, i.e. $m=1$, DDAC-SpAM
takes less time than DAC-SpAM as $m$ increases. The reason is that the
reduced correlation between variables leads to less number of
back-fitting loops required for convergence in the additive model
fitting.  Overall, DDAC-SpAM excels at utilizing all available computing
resources, and is highly efficient and effective compared to existing
algorithms for high-dimensional additive models.

\subsection{Hypothesis Testing}
In this subsection, we investigate the performance of the chi-squared test in
Algorighm~\ref{algo2} with simulated data, under a setting adapted 
from Example 3 in Section~\ref{perf_comp}. The covariates \(X_j\) are generated
in the same way:
\[
    X_j=\frac{W_j+tU_{\lceil j/20\rceil}}{1+t}, j=1,\ldots,p,
\]
where $W_1,\ldots, W_p, U_1,\ldots,U_{\lceil p/20\rceil} \stackrel{i.i.d.}{\sim}$ Uniform $(0, 1)$.
In the model, we include a parameter \(a\) to control the signal-to-noise ratio:
\[
    y_i=a\left[2.5g_1(x_{i1})+2.6g_5(x_{i2})+g_6(x_{i3},2\pi)+g_7(x_{i4},2\pi)\right]+0.5\varepsilon_i.
\]
As before, we fix the sample size $n = 500$, the dimension $p = 10,000$, the truncation size 
$d_n=5$, and the number of machines $m=20$. And we vary the parameter $a$ from 0.1 to 1 in
increments of 0.1. Finally, we set the significance level $\alpha_0=0.05$.

\begin{figure}[t]
  \begin{center}
    \includegraphics[width=12cm]{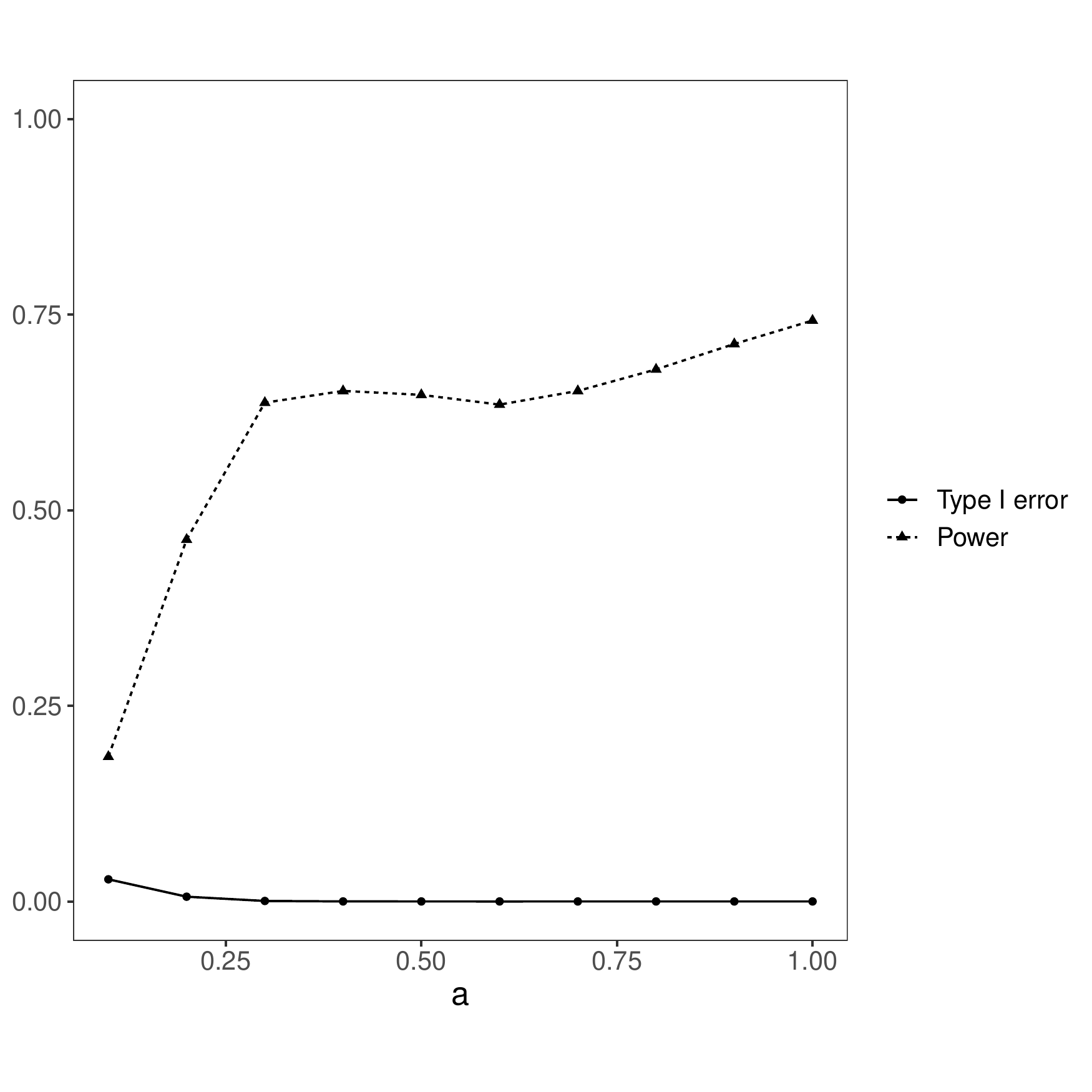}
  \end{center}
  \caption{Type I error and power curves of the chi-squared test in Algorithm~\ref{algo2}. \label{Fig3}}
\end{figure}

Note that due to the lack of formal hypothesis testing algorithms for high-dimensional sparse additive
models, we focus on analyzing the performance of our method only. We report the averages of
\[
    \textrm{Type I error}= (p-4)^{-1}\sum_{j=5}^p 1_{\{\mathcal{T}_{j}>F_{5}(0.95)\}}
\]
and
\[
    \textrm{Power}= 4^{-1} \sum_{j=1}^4 1_{\{\mathcal{T}_{j}>F_{5}(0.95)\}}
\]
over 100 simulation runs in Figure~\ref{Fig3}. The type I error curve stays constantly below the
significance level of 0.05 across the range of $a$, whereas the power curve first shows a steep
positive slope for small $a$ values and then gradually increases as $a$ continues to increase.
The turning point occurs at $a \approx 0.3$, with $SNR \approx 1.3$. 

\section{An Application to Real Data}
\label{sec6}
In this section, we compare the performances of DDAC-SpAM, SpAM, Deco-linear \citep{wang2016decorrelated}, and lasso \citep{tibshirani1996regression}
on the meatspec data set analyzed by \cite{meier2009high} and \cite{YangZougglasso}. The data set was recorded by a Tecator near-infrared spectrometer which
measured the spectrum of light transmitted through a sample of minced pork
meat \citep{thodberg1993ace, borggaard1992optimal}. It is available in the R package \verb"faraway".
Our aim is to predict the fat content by absorbances which can be measured more easily.
This original data set contains $n=215$ observations with $p=100$ predictors
which are highly correlated \citep{meier2009high}. In fact, the average correlation
between different predictors is about 0.986.
After all predictors are centered and scaled to have mean 0 and variance 1, we add
1900 simulated variables from joint distribution $\mathcal{N}(0, \Sigma)$ as artificial noise
terms, where $\Sigma_{ii}=1$ and $\Sigma_{ij}=0.98$ for $i \neq j$, for the purpose of mimicking the high-correlation among the ``actual" predictors. 
Then, DDAC-SpAM, SpAM, Deco-linear, and lasso are applied to predict the fat content using these 2000  features.
We use 10 machines for the DDAC-SpAM algorithm, where the features are distributed randomly. To compare the performances of all methods, we randomly split the dataset into a training set of 172 observations (80\%) and a test set of 43 observations (20\%), which also specifies the truncation size $d_n = \lceil 0.1(172^{1/3}\log 172)\rceil=3$. We repeat the procedure 100 times. 
For each random split, we compute the number of predictors selected, the prediction errors on the test set, and the false positives among the 1900 simulated variables. Table \ref{Realdata1} includes the average values and their associated robust standard deviations over 100 replications.
To evaluate DDAC-SpAM's dependency on the number of local machines, we conduct the same
experiment with varying $m$ values ($m=1,5,10,20,50$), and summarize the results in
Table~\ref{Realdata1-mchange}.

\begin{table}
  \caption{Average prediction error (PE), model size (MS) and false positive count (FP) over 100 repetitions and their robust standard deviations (in parentheses) for DDAC-SpAM, SpAM, Deco-Linear and lasso.  \label{Realdata1}}
  \begin{center}
    \begin{tabular}{lccc}
      Method & PE    & MS & FP\\
      \hline
      DDAC-SpAM & 0.337 (0.087) & 13.84 (4.17) & 7.52 (4.15)\\
      SpAM  & 0.290 (0.084) & 21.30 (5.03) & 9.36 (5.25)\\
      Deco-Linear & 0.453 (0.114) & 58.09 (23.00) & 50.15 (19.14)\\
      Lasso & 0.399 (0.107) & 82.62 (32.99)  & 21.81 (11.18)\\
    \end{tabular}
  \end{center}
\end{table}

\begin{table}
  \caption{Average prediction error (PE), model size (MS) and false positive count (FP) over 100 repetitions and their robust standard deviations (in parentheses) for DDAC-SpAM using $m$ local machines.  \label{Realdata1-mchange}}
  \begin{center}
    \begin{tabular}{lccc}
      $m$ & PE    & MS & FP\\
      \hline
      1 & 0.488 (0.108) & 28.88 (14.12) & 24.38 (13.37)\\
      5  & 0.368 (0.085) & 11.64 (3.53) & 6.40 (3.55)\\
      10 & 0.337 (0.087) & 13.84 (4.17) & 7.52 (4.15)\\
      20 & 0.304 (0.078) & 16.01 (4.50)  & 7.79 (4.38)\\
      50 & 0.292 (0.077) & 15.95 (3.84)  & 6.93 (3.73) \\
    \end{tabular}
  \end{center}
\end{table}

From Table \ref{Realdata1}, DDAC-SpAM outperforms Deco-Linear and lasso in terms of prediction errors. The performance of SpAM is superior to DDAC-SpAM, possibly because it can take advantage of the perfect independence between the original predictors and the 1900 new variables, while the decorrelation step in DDAC-SpAM inadvertently introduces correlation.  Note that DDAC-SpAM selects significantly fewer predictors than competing methods. Considering the high correlation among predictors and to provide a more parsimonious list, DDAC-SpAM could be a very worthwhile method for distributed feature selection. In Table~\ref{Realdata1-mchange}, we observe the prediction errors of DDAC-SpAM steadily decrease as $m$ increases. This is because with more machines used for distributed computing, it is more likely for correlated important features to be separated, making the consistent selection easier.

\section{Discussion}
\label{sec7}
In this paper, we have developed a new feature-distributed learning framework named DDAC-SpAM for the high dimensional additive model. DDAC-SpAM makes predictors less correlated and more suitable for the further sparsistent variable selection. The experiments illustrate that this method not only reduces the computational cost substantially, but also outperforms the existing approach SpAM when covariates are highly correlated.
This is the first work to combine the divide and conquer method with the high dimensional nonparametric model for feature-distributed learning. The results demonstrate that DDAC-SpAM is appealing through the lens of theoretical analysis, empirical performance and its straightforward implemention.

Given that we specifically approximate the additive components by truncated B-spline bases and then impose the sparsity penalty only, DDAC-SpAM framework is readily available for other smoothing method with the additive models, for example, smoothing splines \citep{speckman1985spline} and sparsity-smoothness penalized approaches \citep{meier2009high}.
Besides, extension to the generalized additive model can be an interesting topic for future research.
Lastly, although DDAC-SpAM is currently designed to solve large-$p$-small-$n$ problems, it can be naturally combined with a sample space partition step to deal with large-$p$-large-$n$ problems. The details can be explored in future work.

\section*{Supplementary Material}
The supplementary material consists of Lemma S.1--S.6 and the proofs
of all lemmas, theorems and corollaries.
\section*{Acknowledgements}
We thank the editor, the AE, and anonymous reviewers for their insightful comments which have greatly improved the scope and quality of the paper. \if1\blind
{Zhou was supported by the State Key Program of National Natural Science Foundation of China (71931004) and National Natural Science Foundation of China (92046005) and the National Key R\&D Program of China (2021YFA1000100, 2021YFA1000101). Feng was supported by NIH grant 1R21AG074205-01, NYU University Research Challenge Fund, a grant from NYU School of Global Public Health, and through the NYU IT High Performance Computing resources, services, and staff expertise.} \fi
\section*{Disclosure Statement}
The authors report there are no competing interests to declare.

\putbib

\end{bibunit}

\pagebreak
\vspace{1cm}
\begin{center}
  {\LARGE Supplementary Material for ``DDAC-SpAM: A
    Distributed Algorithm for Fitting High-dimensional Sparse Additive
    Models with Feature Division and Decorrelation"}
\end{center}
\vspace{1cm}

\def\spacingset#1{\renewcommand{\baselinestretch}%
  {#1}\small\normalsize} \spacingset{1}

\spacingset{1.9} %

\begin{bibunit}
In this supplementary material, we will first reproduce Theorem C.1 of
\cite{jia2015preconditioning} for readers' convenience. Next, we will
introduce and prove Lemma S.2--S.6. Last, we will prove Theorem 1,
Corollary 1, and Theorem 2 in the main paper.

\setcounter{lemma}{0}

\renewcommand{\thelemma}{S.\arabic{lemma}}
\begin{lemma}
  \label{VTV}
  \citep{jia2015preconditioning}
  Suppose that $V\in \mathbb{R}^{n\times p}$ comes uniformly from stiefel-manifold. Let $V_B \in \mathbb{R}^{n\times b} $ be any $b$ column of $V$. Suppose that $p-b\geq n$. For any $v_1$, $v_2$, $v_3>0$ with $\sqrt{\frac{b}{n}}+v_1<1 $, $\sqrt{\frac{b}{p}}+v_2 <1$ and $\sqrt{\frac{b}{p}}+v_3 <\frac{1}{3}$, we have
  $$\rP\left[\left\|\frac{p}{n}V_B^{\T}V_B-I_b\right\|\geq \frac{3(\sqrt{\frac{b}{n}}+v_1 )+3(\sqrt{\frac{b}{p}}+v_2)}{1-3(\sqrt{\frac{b}{p}}+v_3 )}\right]\leq 2\exp\{-\frac{nv_1^2}{2}\}+2\exp\{-\frac{pv_2^2}{2}\}+\exp\{-\frac{pv_3^2}{2}\}.$$
\end{lemma}

\renewcommand{\thelemma}{S.\arabic{lemma}}
\begin{lemma}
  Suppose $(p-s)d_n\geq n$. Under Condition 2, for any $t_1$, $t_2$, $t_3>0$ with $\sqrt{\frac{sd_n}{n}}+t_1<1 $, $\sqrt{\frac{s}{p}}+t_2 <1$ and $\sqrt{\frac{s}{p}}+t_3 <\frac{1}{3}$, we have
  \begin{equation*}
    \begin{aligned}
      &\rP\left\{\left\|\frac{pd_n\tildePsiT\tildePsino}{n}\right\|\geq \frac{3(\sqrt{\frac{sd_n}{n}}+t_1 )+3(\sqrt{\frac{s}{p}}+t_2)}{1-3(\sqrt{\frac{s}{p}}+t_3 )}\right\}\\
      \leq & 2\exp\{-\frac{nt_1^2}{2}\}+2\exp\{-\frac{pd_nt_2^2}{2}\}+\exp\{-\frac{pd_nt_3^2}{2}\}.
    \end{aligned}
  \end{equation*}
\end{lemma}

\noindent
\begin{myproof}[Proof of Lemma S.2]
  Let $H=[\tildePsi,\tildePsino]$ and
  $$\eta_1=\left\{\left\|\frac{pd_n}{n} H^{\T}H-I_{sd_n}\right\|\leq \frac{3(\sqrt{\frac{sd_n}{n}}+t_1 )+3(\sqrt{\frac{s}{p}}+t_2)}{1-3(\sqrt{\frac{s}{p}}+t_3) }\right\}.$$
  From Lemma \ref{VTV}, we have
  that
  \begin{gather*}
    \rP\left\{\eta_1 \right\} \geq 1 - \left(2\exp\{-\frac{nt_1^2}{2}\}+2\exp\{-\frac{pd_nt_2^2}{2}\}+\exp\{-\frac{pd_nt_3^2}{2}\}\right),
  \end{gather*}
  for any $t_1$, $t_2$, $t_3>0$ with $\sqrt{\frac{sd_n}{n}}+t_1 <1$, $\sqrt{\frac{s}{p}}+t_2 <1$ and $\sqrt{\frac{s}{p}}+t_3 <\frac{1}{3}$.

  Considering $$H^{\T}H=\left(\begin{matrix}\tildePsiT \tildePsi & \tildePsiT \tildePsino \\ \tildePsinoT \tildePsi &\tildePsinoT \tildePsino \end{matrix} \right)$$
  and
  $$\frac{pd_n}{n}\tildePsiT \tildePsino=\left(\begin{matrix}I_{s_i d_n}& 0_{s_id_n\times (s-s_i)d_n}\end{matrix} \right)\left(\frac{pd_n}{n}H^{\T}H-I_{s d_n}\right)\left(\begin{matrix}0_{s_id_n\times (s-s_i)d_n} \\ I_{(s-s_i)d_n} \end{matrix} \right),$$
  we have
  $$\left\|\frac{pd_n\tildePsiT\tildePsino}{n}\right\|\leq \frac{3(\sqrt{\frac{sd_n}{n}}+t_1 )+3(\sqrt{\frac{s}{p}}+t_2)}{1-3(\sqrt{\frac{s}{p}}+t_3 )},$$
  under the event $\eta_1$.
\end{myproof}

Taking
$t_1=\frac{\tilde\lambda_n}{60C\sqrt{s}}-\sqrt{\frac{sd_n}{n}}>\frac{\tilde\lambda_n}{120C\sqrt{s}}$
(by \(\tilde\lambda_n \to 0\) and
\(\tilde \lambda_n^{-2}n^{-1}s^2d_n \to 0\) in Condition 4),
$t_2=\frac{\tilde\lambda_n}{60C\sqrt{s}}-\sqrt{\frac{s}{p}}$,
$t_3=\frac{1}{10}-\sqrt{\frac{s}{p}} $, we have
$$\rP\left\{\left\|\frac{pd_n\tildePsiT\tildePsino}{n}\right\|\geq \frac{\tilde\lambda_n}{7C\sqrt{s}}\right\}\leq 5\exp\{-\frac{n\tilde\lambda_n^2}{28800C^2s}\}\rightarrow 0.$$

\renewcommand{\thelemma}{S.\arabic{lemma}}
\begin{lemma}
  Suppose $(p-s)d_n\geq n$. Under Condition 2, for any $t_4$, $t_5$, $t_6>0$ with $\sqrt{\frac{sd_n}{n}}+t_4 <1$, $\sqrt{\frac{s}{p}}+t_5 <1$ and $\sqrt{\frac{s}{p}}+t_6 <\frac{1}{3}$, we have
  \begin{equation*}
    \begin{aligned}
      &\rP\left\{\left\| \frac{\sqrt{pd_n}}{\sqrt{n}} \tildePsi \right\|\geq \left[1 + \frac{3(\sqrt{\frac{sd_n}{n}}+t_4 )+3(\sqrt{\frac{s}{p}}+t_5)}{1-3(\sqrt{\frac{s}{p}}+t_6)}\right]^{\frac{1}{2}}\right\}\\
      \leq & 2\exp\{-\frac{nt_4^2}{2}\}+2\exp\{-\frac{pd_nt_5^2}{2}\}+\exp\{-\frac{pd_nt_6^2}{2}\}, 
    \end{aligned}
  \end{equation*}
  and
  \begin{equation*}
    \begin{aligned}
      &\rP\left\{\left\|(\frac{pd_n}{n} \tildePsiT \tildePsi )^{-1}\right\|\geq \left[1-\frac{3(\sqrt{\frac{sd_n}{n}}+t_4 )+3(\sqrt{\frac{s}{p}}+t_5)}{1-3(\sqrt{\frac{s}{p}}+t_6)}\right]^{-1}\right\}\\
      \leq & 2\exp\{-\frac{nt_4^2}{2}\}+2\exp\{-\frac{pd_nt_5^2}{2}\}+\exp\{-\frac{pd_nt_6^2}{2}\}.
    \end{aligned}
  \end{equation*}
\end{lemma}
\noindent
\begin{myproof}[Proof of Lemma S.3]
  Let
  $$\eta_2=\left\{\left\|\frac{pd_n}{n} \widetilde \Psi_S^{T} \widetilde \Psi_S
      -I_{sd_n}\right\|\leq \frac{3(\sqrt{\frac{sd_n}{n}}+t_4
      )+3(\sqrt{\frac{s}{p}}+t_5)}{1-3(\sqrt{\frac{s}{p}}+t_6
      )}\right\}.$$ Then by Lemma S.1, we have
  $$\rP(\eta_2)\geq 1-\left(2\exp\{-\frac{nt_4^2}{2}\}+2\exp\{-\frac{pd_n t_5^2}{2}\}+\exp\{-\frac{pd_n t_6^2}{2}\}\right).$$
  As a submatrix of
  \(\frac{pd_n}{n} \widetilde{\Psi}_S^{T} \widetilde \Psi_S
  -I_{sd_n}\), \(\frac{pd_n}{n} \tildePsiT \tildePsi -I_{s_i d_n}\)
  has the same bound on its norm under event \(\eta_2\). By Weyl's
  inequality for singular values, we have
  \[
    \left\| \frac{\sqrt{pd_n}}{\sqrt{n}} \tildePsi \right\|\leq
    \left[1 + \frac{3(\sqrt{\frac{sd_n}{n}}+t_4
        )+3(\sqrt{\frac{s}{p}}+t_5)}{1-3(\sqrt{\frac{s}{p}}+t_6)}\right]^{\frac{1}{2}}
  \]
  and
  \[
    \left\|(\frac{pd_n}{n} \tildePsiT \tildePsi )^{-1}\right\|\leq \left[1-\frac{3(\sqrt{\frac{sd_n}{n}}+t_4 )+3(\sqrt{\frac{s}{p}}+t_5)}{1-3(\sqrt{\frac{s}{p}}+t_6)}\right]^{-1}.
  \]
\end{myproof}

Taking $t_4=\frac{1}{60}-\sqrt{\frac{sd_n}{n}}>\frac{1}{120}$,
\(t_5=\frac{1}{60}-\sqrt{\frac{s}{p}}\),
$t_6=\frac{1}{10}-\sqrt{\frac{s}{p}}$, we have
\[
  \rP\left\{\left\| \frac{\sqrt{pd_n}}{\sqrt{n}} \tildePsi \right\|
    \geq \sqrt{\frac{8}{7}}\right\} \leq 5\exp\{-\frac{n}{28800}\}\rightarrow 0, 
\]
and
\[
  \rP\left\{\left\|(\frac{pd_n}{n} \tildePsiT \tildePsi
      )^{-1}\right\|\geq \frac{7}{6} \right\} \leq
  5\exp\{-\frac{n}{28800}\}\rightarrow 0.
\]

\renewcommand{\thelemma}{S.\arabic{lemma}}
\begin{lemma}
  Suppose $(p-s_i-1)d_n\geq n$. Under Condition 2, for any $t_8$, $t_9$, $t_{10}>0$ with $\sqrt{\frac{(s_i+1)d_n}{n}}+t_8 <1$, $\sqrt{\frac{s_i+1}{p}}+t_9 <1$ and $\sqrt{\frac{s_i+1}{p}}+t_{10} <\frac{1}{3}$,
  we have
  \begin{equation*}
    \begin{aligned}
      &\rP\left\{\max_{j\in S^{c(i)}}\left\|\frac{pd_n\tildePsiT\tildePsij}{n}\right\|\geq \frac{3(\sqrt{\frac{(s_i+1)d_n}{n}}+t_8 )+3(\sqrt{\frac{s_i+1}{p}}+t_9)}{1-3(\sqrt{\frac{s_i+1}{p}}+t_{10}) }\right\}\\
      \leq & (p_i-s_i)\left(2\exp\{-\frac{nt_8^2}{2}\}+2\exp\{-\frac{pd_nt_9^2}{2}\}+\exp\{-\frac{pd_nt_{10}^2}{2}\}\right).
    \end{aligned}
  \end{equation*}
\end{lemma}

\noindent
\begin{myproof}[Proof of Lemma S.4]
  First, let $G=[\tildePsi,\tildePsij]$, $k\in S^{c(i)}$, and
  \[
    \eta_4=\left\{\max_{j \in S^{c(i)}}\left\|\frac{pd_n}{n}
        G^{\T}G-I_{(s_i+1)d_n}\right\|\leq
      \frac{3(\sqrt{\frac{(s_i+1)d_n}{n}}+t_8
        )+3(\sqrt{\frac{s_i+1}{p}}+t_9)}{1-3(\sqrt{\frac{s_i+1}{p}}+t_{10})
      }\right\}.
  \]
  From Lemma \ref{VTV} and the union bound, we have that
  \begin{equation*}
    \rP(\eta_4) \geq 1 - (p_i - s_i)\left(2\exp\{-\frac{nt_8^2}{2}\}+2\exp\{-\frac{pd_nt_9^2}{2}\}+\exp\{-\frac{pd_nt_{10}^2}{2}\}\right),
  \end{equation*}
  for any $t_8$, $t_9$, $t_{10}>0$ with $\sqrt{\frac{(s_i+1)d_n}{n}}+t_8 <1$, $\sqrt{\frac{s_i+1}{p}}+t_9 <1$ and $\sqrt{\frac{s_i+1}{p}}+t_{10} <\frac{1}{3}$.

  Then, considering $$G^{\T}G=\left(\begin{matrix}\tildePsiT \tildePsi & \tildePsiT \tildePsij \\ \tildePsijT \tildePsi &\tildePsijT \tildePsij \end{matrix} \right)$$
  and
  $$\frac{pd_n}{n}\tildePsiT \tildePsij=\left(\begin{matrix}I_{s_i d_n}& 0_{s_id_n\times d_n}\end{matrix} \right)\left(\frac{pd_n}{n}G^{\T}G-I_{(s_i+1) d_n}\right)\left(\begin{matrix}0_{s_id_n\times d_n} \\ I_{d_n} \end{matrix} \right),$$
  we have
  $$\max_{j\in
    S^{c(i)}}\left\|\frac{pd_n\tildePsiT\tildePsij}{n}\right\|\leq
  \frac{3(\sqrt{\frac{(s_i+1)d_n}{n}}+t_8
    )+3(\sqrt{\frac{s_i+1}{p}}+t_9)}{1-3(\sqrt{\frac{s_i+1}{p}}+t_{10})
  }$$ under the event \(\eta_4\).
\end{myproof}

Taking
$t_{8}=\frac{1}{60\sqrt{s}}-\sqrt{\frac{(s_i+1)d_n}{n}}>\frac{1}{120\sqrt{s}}$
(by \(\tilde\lambda_n \to 0\) and
\(\tilde \lambda_n^{-2}n^{-1}s^2d_n \to 0\) in Condition 4),
$t_{9}=\frac{1}{60\sqrt{s}}-\sqrt{\frac{s_i+1}{p}}$,
$t_{10}=\frac{1}{10}-\sqrt{\frac{s_i+1}{p}}$, we can derive that
\begin{equation*}
  \begin{aligned}
    \rP\left\{\max_{j\in S^{c(i)}}\left\|\frac{pd_n\tildePsiT\tildePsij}{n}\right\|\geq \frac{1}{7{\sqrt{s}}}\right\}
    \leq  5(p_i-s_i)\exp\{-\frac{n}{28800s}\}\rightarrow 0.
  \end{aligned}
\end{equation*}

\renewcommand{\thelemma}{S.\arabic{lemma}}
\begin{lemma}
  Suppose $(p-1)d_n\geq n$. Under Condition 2, for any $t_{11}$, $t_{12}$, $t_{13}>0$ with $\sqrt{\frac{d_n}{n}}+t_{11} <1$, $\sqrt{\frac{1}{p}}+t_{12} <1$ and $\sqrt{\frac{1}{p}}+t_{13} <\frac{1}{3}$,
  we have
  \begin{equation*}
    \begin{aligned}
      &\rP\left\{\max_{j \in S^{c(i)}}\lef \frac{\sqrt{pd_n}}{\sqrt{n}}\tildePsij \rig \geq \left[1+\frac{3(\sqrt{\frac{d_n}{n}}+t_{11} )+3(\sqrt{\frac{1}{p}}+t_{12})}{1-3(\sqrt{\frac{1}{p}}+t_{13} )}\right]^{\frac{1}{2}}\right\}\\
      \leq & (p_i - s_i)\left(2\exp\{-\frac{nt_{11}^2}{2}\}+2\exp\{-\frac{pd_nt_{12}^2}{2}\}+\exp\{-\frac{pd_nt_{13}^2}{2}\}\right), 
    \end{aligned}
  \end{equation*}
  and
  \begin{equation*}
    \begin{aligned}
      &\rP\left\{\max_{j \in S^{c(i)}} \lef (\frac{pd_n}n \tildePsijT \tildePsij)^{-1} \rig \geq \left[1 - \frac{3(\sqrt{\frac{d_n}{n}}+t_{11})+3(\sqrt{\frac{1}{p}}+t_{12})}{1-3(\sqrt{\frac{1}{p}}+t_{13})}\right]^{-1}\right\}\\
      \leq & (p_i - s_i)\left(2\exp\{-\frac{nt_{11}^2}{2}\}+2\exp\{-\frac{pd_nt_{12}^2}{2}\}+\exp\{-\frac{pd_nt_{13}^2}{2}\}\right).
    \end{aligned}
  \end{equation*}
\end{lemma}

\noindent
\begin{myproof}[Proof of Lemma S.5]
  Let
  \begin{equation*}
    \begin{aligned}
      \eta_5=\left\{\max_{j \in S^{c(i)}}\left\|(\frac{pd_n}{n} \tildePsijT \tildePsij )-I_{d_n}\right\|\leq \frac{3(\sqrt{\frac{d_n}{n}}+t_{11} )+3(\sqrt{\frac{1}{p}}+t_{12})}{1-3(\sqrt{\frac{1}{p}}+t_{13} )}\right\}.
    \end{aligned}
  \end{equation*}
  Then by Lemma S.1 and the union bound, we have
  $$\rP(\eta_5)\geq
  1-(p_i - s_i)\left(2\exp\{-\frac{nt_{11}^2}{2}\}-2\exp\{-\frac{pd_nt_{12}^2}{2}\}-\exp\{-\frac{pd_nt_{13}^2}{2}\}\right).$$
  By Weyl's inequality for singular values, we have
  \[
    \max_{j \in S^{c(i)}}\lef \frac{\sqrt{pd_n}}{\sqrt{n}}\tildePsij \rig \leq \left[1+\frac{3(\sqrt{\frac{d_n}{n}}+t_{11} )+3(\sqrt{\frac{1}{p}}+t_{12})}{1-3(\sqrt{\frac{1}{p}}+t_{13} )}\right]^{\frac{1}{2}}
  \]
  and
  \[
    \max_{j \in S^{c(i)}} \lef (\frac{pd_n}n \tildePsijT
    \tildePsij)^{-1} \rig \leq \left[1 -
      \frac{3(\sqrt{\frac{d_n}{n}}+t_{11}
        )+3(\sqrt{\frac{1}{p}}+t_{12})}{1-3(\sqrt{\frac{1}{p}}+t_{13}
        )}\right]^{-1}
  \]
  under the event \(\eta_5\).
\end{myproof}

Taking $t_{11}=\frac{1}{60}-\sqrt{\frac{d_n}{n}} > \frac{1}{120}$, $t_{12}=\frac{1}{60}-\sqrt{\frac{1}{p}}$, $t_{13}=\frac{1}{10}-\sqrt{\frac{1}{p}}$, it can be derived that
\[
  \rP\left\{\max_{j \in S^{c(i)}}\lef
    \frac{\sqrt{pd_n}}{\sqrt{n}}\tildePsij \rig \geq
    \sqrt{\frac{8}{7}}\right\}\leq 5(p_i - s_i)\exp\{-\frac{n}{28800}\}\rightarrow 0, 
\]
and
\[
  \rP\left\{\max_{j \in S^{c(i)}} \lef (\frac{pd_n}n \tildePsijT
    \tildePsij)^{-1} \rig \geq \frac{7}{6} \right\} \leq 5(p_i -
  s_i)\exp\{-\frac{n}{28800}\}\rightarrow 0.
\]

\renewcommand{\thelemma}{S.\arabic{lemma}}
\begin{lemma}
  Suppose $(p-s+s_i-1)d_n\geq n$. Under Condition 2, for any $t_{14}$, $t_{15}$, $t_{16}>0$ with $\sqrt{\frac{(s-s_i+1)d_n}{n}}+t_{14} <1$, $\sqrt{\frac{s-s_i+1}{p}}+t_{15} <1$ and $\sqrt{\frac{s-s_i+1}{p}}+t_{16} <\frac{1}{3}$,
  we have
  \begin{equation*}
    \begin{aligned}
      &\rP\left\{\max_{j \in S^{c(i)}}\left\|\frac{pd_n\tildePsinoT\tildePsij}{n}\right\|\geq \frac{3(\sqrt{\frac{(s-s_i+1)d_n}{n}}+t_{14} )+3(\sqrt{\frac{s-s_i+1}{p}}+t_{15})}{1-3(\sqrt{\frac{s-s_i+1}{p}}+t_{16})}\right\}\\
      \leq & (p_i - s_i)\left(2\exp\{-\frac{nt_{14}^2}{2}\}+2\exp\{-\frac{pd_nt_{15}^2}{2}\}+\exp\{-\frac{pd_nt_{16}^2}{2}\}\right).
    \end{aligned}
  \end{equation*}
\end{lemma}
\noindent
\begin{myproof}[Proof of Lemma S.6]
  Let $K=[\tildePsino,\tildePsij]$ for \(k \in S^{c(i)}\) and
  \[
    \eta_6 = \left\{\max_{j \in S^{c(i)}}\left\|\frac{pd_n}{n}
        K^{\T}K-I_{(s-s_i+1)d_n}\right\|\leq
      \frac{3(\sqrt{\frac{(s-s_i+1)d_n}{n}}+t_{14}
        )+3(\sqrt{\frac{s-s_i+1}{p}}+t_{15})}{1-3(\sqrt{\frac{s-s_i+1}{p}}+t_{16})
      }\right\}.
  \]
  From Lemma S.1 and the union bound, we have that
  \begin{equation*}
    \begin{aligned}
      \rP(\eta_6) \geq 1 - (p_i - s_i)\left(2\exp\{-\frac{nt_{14}^2}{2}\}+2\exp\{-\frac{pd_nt_{15}^2}{2}\}+\exp\{-\frac{pd_nt_{16}^2}{2}\}\right),
    \end{aligned}
  \end{equation*}
  for any $t_{14}$, $t_{15}$, $t_{16}>0$ with $\sqrt{\frac{(s-s_i+1)d_n}{n}}+t_{14} <1$, $\sqrt{\frac{s-s_i+1}{p}}+t_{15} <1$ and $\sqrt{\frac{s-s_i+1}{p}}+t_{16} <\frac{1}{3}$.

  Considering $$K^{\T}K=\left(\begin{matrix}\tildePsinoT \tildePsino & \tildePsinoT \tildePsij \\ \tildePsijT \tildePsino &\tildePsijT \tildePsij \end{matrix} \right)$$
  and
  $$\frac{pd_n}{n}\tildePsinoT \tildePsij=\left(\begin{matrix}I_{(s-s_i) d_n}& 0_{(s-s_i)d_n\times d_n}\end{matrix} \right)\left(\frac{pd_n}{n}K^{\T}K-I_{(s-s_i+1) d_n}\right)\left(\begin{matrix}0_{(s-s_i)d_n\times d_n} \\ I_{d_n} \end{matrix} \right),$$
  we have
  \[
    \max_{j \in
      S^{c(i)}}\left\|\frac{pd_n\tildePsinoT\tildePsij}{n}\right\|\leq
    \frac{3(\sqrt{\frac{(s-s_i+1)d_n}{n}}+t_{14}
      )+3(\sqrt{\frac{s-s_i+1}{p}}+t_{15})}{1-3(\sqrt{\frac{s-s_i+1}{p}}+t_{16})}
  \]
  under the event \(\eta_6\).
\end{myproof}
Taking
$t_{14}=\frac{\tilde\lambda_n}{60\sqrt{s}C}-\sqrt{\frac{(s-s_i+1)d_n}{n}}
> \frac{\tilde\lambda_n}{120\sqrt{s}C}$ (by
\(\tilde\lambda_n \to 0\) and
\(\tilde \lambda_n^{-2}n^{-1}s^2d_n \to 0\) in Condition 4),
$t_{15}=\frac{\tilde\lambda_n}{60\sqrt{s}C}-\sqrt{\frac{s-s_i+1}{p}}$,
$t_{16}=\frac{1}{10}-\sqrt{\frac{s-s_i+1}{p}}$, it can be derived that
\[
  \rP\left\{\max_{j \in
      S^{c(i)}}\left\|\frac{pd_n\tildePsinoT\tildePsij}{n}\right\|\geq
    \frac{\tilde\lambda_n}{7 C\sqrt{s}}\right\} \leq 5(p_i -
  s_i)\exp\{-\frac{n\tilde\lambda_n^2}{28800C^2s}\} \to 0.
\]

For convenience, let $\beta^*$ denote the true coefficient vector $\beta$ in (4) for the rest of this supplementary material.
\begin{myproof}[Proof of Theorem 1]
  For the $i$-th group, we have
  \setcounter{equation}{0}
  \begin{equation}
    \renewcommand\theequation{S.\arabic{equation}}
    \tildeY=\tildePsi \beta_{S}^{*(i)}+\widetilde{Z}+W,
    \label{Ey}
  \end{equation}
  where
  $W=\tildePsino \beta_{S}^{*(-i)}+\tilde{\varepsilon}=W_{1}+W_{2}$.

  A vector $\hat{\beta}^{(i)} \in \mathbb{R}^{d_{n} p_{i}}$ is the minimizer of the objective function 
  \begin{equation}
    \renewcommand\theequation{S.\arabic{equation}}
    \begin{aligned}
      &R_{n}\left(\beta^{(i)}\right)+\lambda_{n} \Omega\left(\beta^{(i)}\right)\\
      =&\frac{1}{2 n}\left\|\tildeY-\sum_{k=1}^{p_ i} \tildePsij \beta_{k}^{(i)}\right\|^{2}+\lambda_n\sum_{k=1}^{p_i}\sqrt{\onen\beta_{k}^{(i)\T}\tildePsijT \tildePsij\beta_{k}^{(i)}}\\
      =&\frac{1}{2 n}\left\|\tildeY-\sum_{k=1}^{p_i} \tildePsij \beta_{k}^{(i)}\right\|^{2}+\tilde\lambda_n \sum_{k=1}^{p_ i}\left\|\frac{1}{\sqrt{npd_n}} \tildePsij\beta_{k}^{(i)}\right\|
    \end{aligned}
    \label{Ecompose}
  \end{equation}
  if and only if there exists a subgradient $\hat{g}^{(i)} \in \partial \Omega\left(\hat{\beta}^{(i)}\right)$, such that
  \begin{equation}
    \renewcommand\theequation{S.\arabic{equation}}
    \frac{1}{n} \widetilde{\Psi}^{(i) \T}\left(\sum_{j=1}^{p_i} \widetilde{\Psi}_{j}^{(i)} \hat{\beta}_{j}^{(i)}-\widetilde{Y}\right)+\tilde\lambda_{n} \hat{g}^{(i)}=0.
    \label{Estationary}
  \end{equation}
  The subdifferential \(\partial \Omega ( \beta^{(i)} )\) is the set of vectors \(g^{(i)} \in \mathbb{R}^{p_{i} d_{n}}\) satisfying
  \begin{gather*}
    g_{k}^{(i)}=\frac{\frac{1}{npd_n}\widetilde{\Psi}_{k}^{(i)\T} \widetilde{\Psi}_{k}^{(i)}\beta_{k}^{(i)} }{\sqrt{\frac{1}{npd_n}\beta_{k}^{(i)\T}\widetilde{\Psi}_{k}^{(i)\T} \widetilde{\Psi}_{k}^{(i)}\beta_{k}^{(i)}}}, \quad \textrm{if} \quad \beta_{k}^{(i)} \neq 0,\\
    g_{k}^{(i)\T}(\frac{1}{npd_n}\widetilde{\Psi}_{k}^{(i) \T} \widetilde{\Psi}_{k}^{(i)})^{-1}g_{k}^{(i)}\leq 1 , \quad \textrm{if} \quad \beta_{k}^{(i)} = 0.
  \end{gather*}

  We use ``witness" proof techniques \citep{wainwright2006sharp}, i.e., set $\hat{\beta}_{S^{c}}^{(i)}=0$ and $\hat{g}_{S}^{(i)}=\partial \Omega\left(\beta^{*(i) }\right)_{S}$.  We then obtain $\hat{\beta}_{S}^{(i)}$ and $\hat{g}_{S^ c}^{(i)}$ from the stationary condition in (S.\ref{Estationary}). By showing that, with high probability
  $\hat{\beta}_{k}^{(i)} \neq 0 $ for $k\in S$ and $g_{k}^{(i)\T}(\frac{1}{npd_n}\widetilde{\Psi}_{k}^{(i) \T} \widetilde{\Psi}_{k}^{(i)})^{-1}g_{k}^{(i)}\leq 1 $ for $k\in S^c$, we can then demonstrate that with high probability there exists a minimizer to the optimization problem in (S.\ref{Ecompose}) that has the same sparsity pattern as the true model.

  Setting $\hat{\beta}_{S^{c}}^{(i)}=0$ and
  $\hat{g}_{k}^{(i)}=\frac{\frac1{npd_n} \widetilde{\Psi}_{k}^{(i)\T}
    \widetilde{\Psi}_{k}^{(i)}\beta_{k}^{(i)}
  }{\sqrt{\frac{1}{npd_n}\beta_{k}^{(i)\T}\widetilde{\Psi}_{k}^{(i)\T}
      \widetilde{\Psi}_{k}^{(i)}\beta_{k}^{(i)}}}$ for $k\in S^{(i)}$,
  the stationary condition for $\hat\beta_S^{(i)}$ is
  \begin{equation*}
    \frac{1}{n}\tildePsiT(\tildePsi \hat\beta_S^{(i)} - \tildeY)+\tilde\lambda_n \hatgSi=0.
  \end{equation*}

  With (S.\ref{Ey}), it can be written as
  $$\frac{1}{n}\tildePsiT\widetilde{\Psi}_{S}^{(i)}\left(\hat\beta_{S}^{(i)}-\beta_{S}^{*(i)}\right)-\onen \tildePsiT W_1-\onen \tildePsiT W_2-\onen \tildePsiT \widetilde{Z}+\tilde\lambda_n\hatgSi=0$$
  or
  \begin{equation*}
    \begin{aligned}
      \hat\beta_{S}^{(i)}-\betaSti=(\onen\tildePsiT \tildePsi)^{-1}(\onen \tildePsiT W_1+\onen \tildePsiT W_2+\onen \tildePsiT \widetilde{Z} - \tilde\lambda_n\hatgSi),
    \end{aligned}
  \end{equation*}
  assuming that $\onen\tildePsiT \tildePsi$ is nonsingular.

  Recalling our definition $\rho_n=\min_{k\in S^{(i)}}\left\| \beta_k^{*(i)}\right\|_\infty > 0$, it suffices to show that
  \begin{equation}
    \renewcommand\theequation{S.\arabic{equation}}
    \left\|\hat\beta_{S}^{(i)}-\beta_{S}^{*(i) }\right\|_{\infty}<\frac{\rho_{n}}{2}
    \label{betabound}
  \end{equation}
  in order to ensure that $\textrm{supp}(\hat\beta_{S}^{(i)})=\textrm{supp}(\betaSti)=\{k:\left\|\beta_{k}^{*(i)}\right\|_{\infty} \neq 0\}$.

  Using $\SigmaSS=\onen(\tildePsiT\tildePsi)$ to simplify notation, we have the $l_\infty$ bound:
  \begin{equation}
    \renewcommand\theequation{S.\arabic{equation}}
    \begin{aligned}
      \left\|\hat\beta_{S}^{(i)}-\beta_{S}^{*(i) }\right\|_{\infty} \leq & \left\|\hat\beta_{S}^{(i)}-\beta_{S}^{*(i) }\right\| \leq \left\|\SigmaSSinv(\onen\tildePsiT W_1)\right\|\\ & +\left\|\SigmaSSinv(\onen\tildePsiT W_2)\right\| +\left\|\SigmaSSinv(\onen\tildePsiT \widetilde{Z})\right\|+\tilde\lambda_n\left\|\SigmaSSinv\hatgSi\right\|.
    \end{aligned}
    \label{betaerror}
  \end{equation}

  Now, we proceed to bound the first term of (S.\ref{betaerror}).
  Notice that derived from Condition 1,
  $\left\|\beta_{S}^{*(-i)}\right\|\leq \sqrt{s-s_i}C \leq \sqrt{s}C$,
  then
  \begin{equation}
    \renewcommand\theequation{S.\arabic{equation}}
    \begin{aligned}
      & \left\|\SigmaSSinv(\onen \tildePsiT W_1 )\right\| \leq \left\|\SigmaSSinv \right\|\left\|\onen \tildePsiT \tildePsino\right\|\left\|\betaStno\right\|\\
      \leq &C\left[1-\frac{3(\sqrt{\frac{sd_n}{n}}+t_4 )+3(\sqrt{\frac{s}{p}}+t_5)}{1-3(\sqrt{\frac{s}{p}}+t_6)}\right]^{-1}\left[\frac{3(\sqrt{\frac{sd_n}{n}}+t_1 )+3(\sqrt{\frac{s}{p}}+t_2)}{1-3(\sqrt{\frac{s}{p}}+t_3)}\right]\sqrt{s},
    \end{aligned}
    \label{1.1}
  \end{equation}
  under events \(\eta_1\) and \(\eta_2\), where the last inequality is
  derived with Lemma S.2 and Lemma S.3.
  $t_1, t_2, t_3, t_4, t_5, t_6$ are some positive constants as shown
  in Lemma S.2 and S.3.

  Then, consider the second term
  $\left\|\SigmaSSinv(\onen \tildePsiT W_2)\right\| \leq
  \|\SigmaSSinv\| \| T\e \|$, where
  \(T = \frac{1}{n} \widetilde{\Psi}_{S}^{\T} F\). We first condition
  on \(X\), so that \(T\) becomes a deterministic \(sd_n\) by \(n\)
  matrix. This implies \(T\e | X \sim \mathcal{N}(0, \sigma^2TT^{\T})\), where
  \begin{equation*}
    \|\sigma^2TT^{\T}\|=\frac{\sigma^2}{n^2}\left\|\widetilde{\Psi}_{S}^{\T} FF^{\T}\widetilde{\Psi}_{S}\right\| \leq \frac{\sigma^2}{n^2}\|F\|^2\|\widetilde{\Psi}_{S}\|^2.
  \end{equation*}
  As a result, \(nT \e/(\sigma\|F\|\|\widetilde{\Psi}_{S}\|)\Big|X\)
  follows an \(sd_n\)-dimensional Gaussian distribution whose
  covariance matrix has eigenvalues less then \(1\), so we can apply
  Lemma 1 of \cite{laurent2000adaptive}:
  \[
    \rP\left(n\|T \e\|\Big/\left(\sigma\|F\|
        \|\widetilde{\Psi}_{S}\|\right) \leq \sqrt{5s
        d_n}\Big|X\right) \geq 1 - \exp(-sd_n).
  \]
  Taking expectation with respect to \(X\), this yields
  \[
    \rP\left(\|T \e\| \leq \frac{\sigma}{n}\sqrt{5s
        d_n}\|F\|\|\widetilde{\Psi}_{S}\|\right) \geq 1 - \exp(-sd_n).
  \]
  By Condition 3, we get
  \[
    \rP\left(\|F\| \leq \frac{1}{\sqrt{pd_n\delta n^{\alpha -
            1}}}\right) \geq 1 - \exp(-\xi n^\gamma).
  \]
  Denote the events
  \[
    \eta_3 = \left\{\|T \e\| \leq \sqrt{5s d_n}
      \|F\|\|\widetilde{\Psi}_{S}\|\sigma/n\right\}
  \] and
  \[
    \eta_8 = \left\{\|F\| \leq 1/\sqrt{pd_n\delta n^{\alpha -
          1}}\right\}.
  \]
  Under \(\eta_2\), \(\eta_3\) and \(\eta_8\), we
  have
  \begin{equation}
    \renewcommand\theequation{S.\arabic{equation}}
    \begin{multlined}
      \left\|\SigmaSSinv\left(\onen \tildePsiT W_2\right)\right\| \leq
      \sqrt{5} \sigma \delta^{-\frac{1}{2}}
      \left[1-\frac{3(\sqrt{\frac{sd_n}{n}}+t_4
          )+3(\sqrt{\frac{s}{p}}+t_5)}{1-3(\sqrt{\frac{s}{p}}+t_6)}\right]^{-1}\\
      \left[1 + \frac{3(\sqrt{\frac{sd_n}{n}}+t_4
          )+3(\sqrt{\frac{s}{p}}+t_5)}{1-3(\sqrt{\frac{s}{p}}+t_6)}\right]^{\frac12}
      n^{-\frac{\alpha}2}\sqrt{sd_n}
    \end{multlined}
    \label{1.2}
  \end{equation}

  Then, we need to bound
  $\left\|\SigmaSSinv(\onen \tildePsiT \widetilde{Z})\right\| \leq
  \onen\left\|\SigmaSSinv \tildePsiT \right\| \left\|FZ \right\| \leq
  \sqrt{\onen \left\|\SigmaSSinv \right\|}
  \left\|F\right\|\left\|Z\right\|$.  Working over the Sobolev spaces
  $\mathcal{S}_j$ (Condition 1),
  \begin{equation}
    \renewcommand\theequation{S.\arabic{equation}}
    \begin{aligned}
      \label{zequation}
      |z_i|&=|\sum_{j\in S}\sum_{k=d_n+1}^{\infty}\beta_{jk}^{*}\psi_{jk}(x_{ij})|
             \leq B\sum_{j\in S}\sum_{k=d_n+1}^{\infty}|\beta_{jk}^{\ast}|
             \leq\frac{sC'}{d_n^{3/2}},
    \end{aligned}
  \end{equation}
  for some constant $C'>0$.  Thus, we have
  $\lef Z \rig\leq \frac{\sqrt{n}sC'}{d_n^{3/2}}$. It then can be
  derived that
  \begin{equation}
    \renewcommand\theequation{S.\arabic{equation}}
    \begin{aligned}
      \left\|\SigmaSSinv\left(\onen \tildePsiT \widetilde{Z}\right)\right\| \leq C' \delta^{-\frac{1}{2}} \left[1-\frac{3(\sqrt{\frac{sd_n}{n}}+t_4 )+3(\sqrt{\frac{s}{p}}+t_5)}{1-3(\sqrt{\frac{s}{p}}+t_6)}\right]^{-\frac{1}{2}} n^{\frac{1}{2}(1 - \alpha)}sd_n^{-\frac{3}{2}},
    \end{aligned}
    \label{1.3}
  \end{equation}
  under events \(\eta_2\) and \(\eta_8\).

  Finally, we consider the term $\tilde\lambda_n \lef \SigmaSSinv \hatgSi\rig$.
  Note that for $k\in S^{(i)}$,
  $$
  1= \hat{g}_k^{(i)\T}\left(\frac1{npd_n} \tildePsijT \tildePsij\right)^{-1}\hat{g}^{(i)}_k\geq \frac{pd_n}{\lef \frac1{n} \tildePsijT \tildePsij\rig} \lef\hat{g}^{(i)}_k \rig^2
  $$
  and thus
  $\lef \hat{g}^{(i)}_k \rig\leq \sqrt{\lef \frac1{n} \tildePsijT
    \tildePsij\rig\big/(pd_n)} \leq \sqrt{\lef \SigmaSS
    \rig\big/(pd_n)}$. Therefore,
  $$\tilde\lambda_n\lef \SigmaSSinv \hatgSi \rig \leq \tilde\lambda_n \sqrt{s_i} \lef
  \SigmaSSinv \rig \max_{k \in S^{(i)}}\lef \hat{g}^{(i)}_k \rig\leq
  \tilde\lambda_n \sqrt{s} \lef \frac{\SigmaSSinv}{pd_n}\rig
  \sqrt{pd_n\lef \SigmaSS\rig} .$$ It follows that
  \begin{equation}
    \renewcommand\theequation{S.\arabic{equation}}
    \begin{aligned}
      &\tilde\lambda_n \lef \SigmaSSinv \hatgSi\rig\\
      \leq & \left[1-\frac{3(\sqrt{\frac{sd_n}{n}}+t_4 )+3(\sqrt{\frac{s}{p}}+t_5)}{1-3(\sqrt{\frac{s}{p}}+t_6 )}\right]^{-1}\left[1+\frac{3(\sqrt{\frac{sd_n}{n}}+t_4 )+3(\sqrt{\frac{s}{p}}+t_5)}{1-3(\sqrt{\frac{s}{p}}+t_6 )}\right]^{\frac{1}{2}} \sqrt{s}\tilde\lambda_n,
    \end{aligned}
    \label{1.4}
  \end{equation}
  under event \(\eta_2\).

  Under Condition 1, 2, 3 and 4, we combine results (S.\ref{1.1}),
  (S.\ref{1.2}), (S.\ref{1.3}), (S.\ref{1.4}) with (S.\ref{betaerror})
  and take
  $t_1=\frac{\tilde\lambda_n}{60C\sqrt{s}}-\sqrt{\frac{sd_n}{n}}$,
  $t_2=\frac{\tilde\lambda_n}{60C\sqrt{s}}-\sqrt{\frac{s}{p}}$,
  $t_3=\frac{1}{10}-\sqrt{\frac{s}{p}} $,
  $t_4=\frac{1}{60}-\sqrt{\frac{sd_n}{n}}$,
  $t_5=\frac{1}{60}-\sqrt{\frac{s}{p}}$,
  \(t_6=\frac{1}{10}-\sqrt{\frac{s}{p}}\), which yields for
  sufficiently large \(n\),
  \begin{equation}
    \renewcommand\theequation{S.\arabic{equation}}
    \begin{aligned}
      &\left\|\hat\beta_{S}^{(i)}-\beta_{S}^{*(i) }\right\|_{\infty} \leq \left\|\hat\beta_{S}^{(i)}-\beta_{S}^{*(i) }\right\|\\
      \leq & C\left[1-\frac{3(\sqrt{\frac{sd_n}{n}}+t_4 )+3(\sqrt{\frac{s}{p}}+t_5)}{1-3(\sqrt{\frac{s}{p}}+t_6)}\right]^{-1}\left[\frac{3(\sqrt{\frac{sd_n}{n}}+t_1 )+3(\sqrt{\frac{s}{p}}+t_2)}{1-3(\sqrt{\frac{s}{p}}+t_3)}\right]\sqrt{s}\\
      &+\sqrt{5} \sigma \delta^{-\frac{1}{2}} \left[1-\frac{3(\sqrt{\frac{sd_n}{n}}+t_4 )+3(\sqrt{\frac{s}{p}}+t_5)}{1-3(\sqrt{\frac{s}{p}}+t_6 )}\right]^{-1}\left[1+\frac{3(\sqrt{\frac{sd_n}{n}}+t_4 )+3(\sqrt{\frac{s}{p}}+t_5)}{1-3(\sqrt{\frac{s}{p}}+t_6 )}\right]^{\frac{1}{2}}\sqrt{sd_n}\\
      &+C' \delta^{-\frac{1}{2}} \left[1-\frac{3(\sqrt{\frac{sd_n}{n}}+t_4 )+3(\sqrt{\frac{s}{p}}+t_5)}{1-3(\sqrt{\frac{s}{p}}+t_6)}\right]^{-\frac{1}{2}} n^{\frac{1}{2}(1 - \alpha)}sd_n^{-\frac{3}{2}} \\
      &+\left[1-\frac{3(\sqrt{\frac{sd_n}{n}}+t_4 )+3(\sqrt{\frac{s}{p}}+t_5)}{1-3(\sqrt{\frac{s}{p}}+t_6 )}\right]^{-1}\left[1+\frac{3(\sqrt{\frac{sd_n}{n}}+t_4 )+3(\sqrt{\frac{s}{p}}+t_5)}{1-3(\sqrt{\frac{s}{p}}+t_6 )}\right]^{\frac{1}{2}}\sqrt{s}\tilde\lambda_n\\
      =&\frac{\tilde\lambda_n}{6}+\sqrt{\frac{70}{9}}\sigma\delta^{-1/2}n^{-\alpha/2}\sqrt{sd_n}+\sqrt{\frac{7}{6}}C'\delta^{-1/2}n^{(1 - \alpha)/2}sd_n^{-3/2}+\sqrt{\frac{14}{9}}\sqrt{s}\tilde\lambda_n\\
      \leq&\frac{\rho_{n}}{2}
    \end{aligned}
    \label{1.5}
  \end{equation}
  under events \(\eta_1\) \(\eta_2\) \(\eta_3\) and \(\eta_8\). This
  proves (S.\ref{betabound}).

  Now, we analyze $\hatgSCi$.
  We require that
  $$\hatgjiT\left(\frac1{npd_n} \widetilde{\Psi}_k^{(i)\T}\widetilde{\Psi}_k^{(i)}\right)^{-1}\hatgji < 1, \quad \textrm{for all} \quad k\in S^{c(i)}, $$
  where strict inequality is imposed so that the minimizer of
  (S.\ref{Ecompose}) is unique. Since
  $$\hatgjiT\left(\frac{1}{npd_n}
    \widetilde{\Psi}_k^{(i)\T}\widetilde{\Psi}_k^{(i)}\right)^{-1}\hatgji
  \leq \lef \hatgji \rig^2 \lef \left(\frac{1}{npd_n}
    \widetilde{\Psi}_k^{(i)\T}\widetilde{\Psi}_k^{(i)}\right)^{-1}
  \rig,$$ it suffices to show that
  \begin{equation}
    \renewcommand\theequation{S.\arabic{equation}}
    \max_{k\in S^{c(i)}}\lef pd_n \hatgji \rig \sqrt{\max_{k\in S^{c(i)}}
    \lef \left(\frac{pd_n}{n}
        \widetilde{\Psi}_k^{(i)\T}\widetilde{\Psi}_k^{(i)}\right)^{-1}\rig}
    < 1.
    \label{subgra_bound}
  \end{equation}

  Recall that we have set $\hat{\beta}_{S^c}^{(i)}=\beta_{S^c}^{*(i)}=0$. The stationary condition for $k\in S^{c(i)}$ is thus given by
  $$\onen \tildePsijT \left(\tildePsi\hat{\beta}_S^{(i)}-\tildePsi\beta_S^{*(i)}-\widetilde{Z}-W\right)+\tilde\lambda_n\hatgji=0.$$
  Therefore, for $k\in S^{c(i)}$,
  \begin{equation}
    \renewcommand\theequation{S.\arabic{equation}}
    \begin{aligned}
      &\hatgji=\frac{1}{\tilde\lambda_n}\cdot\onen\tildePsijT\left(\tildePsi\beta_S^{*(i)}-\tildePsi\hat{\beta}_S^{(i)}+\widetilde{Z}+W\right)\\
      &=\frac{1}{\tilde\lambda_n}\cdot\onen\tildePsijT\tildePsi\left(\beta_S^{*(i)}-\hat{\beta}_S^{(i)}\right)
        +\frac{1}{\tilde\lambda_n}\cdot\onen\tildePsijT \widetilde{Z}+\frac{1}{\tilde\lambda_n}\cdot\onen\tildePsijT W_1+\frac{1}{\tilde\lambda_n}\cdot\onen\tildePsijT W_2\\
      &:=\mathcal{G}_1+\mathcal{G}_2+\mathcal{G}_3+\mathcal{G}_4.
    \end{aligned}
    \label{2.1}
  \end{equation}

  By (S.\ref{1.5}) and Lemma S.4, under events $\eta_1$, $\eta_2$,
  \(\eta_3\), \(\eta_4\) and \(\eta_8\), we have
  \begin{equation}
    \renewcommand\theequation{S.\arabic{equation}}
    \begin{aligned}
      \max_{k \in S^{c(i)}}\lef \mathcal{G}_1 \rig &\leq \frac{1}{\tilde\lambda_n} \max_{k \in S^{c(i)}}\lef \onen\tildePsijT\tildePsi \rig \lef \beta_S^{*(i)}-\hat{\beta}_S^{(i)} \rig \\
                                                   &\leq \frac{1}{pd_n\tilde\lambda_n}\cdot \frac{3(\sqrt{\frac{(s_i+1)d_n}{n}}+t_8
                                                     )+3(\sqrt{\frac{s_i+1}{p}}+t_9)}{1-3(\sqrt{\frac{s_i+1}{p}}+t_{10}) }\\
                                                   &\cdot
                                                     \left(\frac{\tilde\lambda_n}{6}+\sqrt{\frac{70}{9}}\sigma\delta^{-1/2}n^{-\alpha/2}\sqrt{sd_n}+\sqrt{\frac{7}{6}}C'\delta^{-1/2}n^{(1 - \alpha)/2}sd_n^{-3/2}+\sqrt{\frac{14}{9}}\sqrt{s}\tilde\lambda_n\right).
    \end{aligned}
    \label{2.2}
  \end{equation}
  
  Meanwhile, by Lemma S.5 and (S.\ref{zequation}), under the event
  \(\eta_5\) we have
  \begin{equation}
    \renewcommand\theequation{S.\arabic{equation}}
    \begin{aligned}
      \max_{k \in S^{c(i)}}\lef \mathcal{G}_2 \rig & =\max_{k \in S^{c(i)}}\lef \frac{1}{\tilde\lambda_n}\cdot\onen\tildePsijT F Z \rig \leq \frac{1}{\tilde\lambda_n\sqrt{n}} \max_{k \in S^{c(i)}}\lef \frac{1}{\sqrt{n}} \tildePsij \rig \lef F \rig \lef Z \rig\\
                                                   &\leq C'(pd_n)^{-1}\delta^{-1/2}\tilde\lambda_n^{-1}n^{(1 - \alpha)/2}sd_n^{-3/2}
                                                     \left[1+\frac{3(\sqrt{\frac{d_n}{n}}+t_{11} )+3(\sqrt{\frac{1}{p}}+t_{12})}{1-3(\sqrt{\frac{1}{p}}+t_{13} )}\right]^{\frac{1}{2}}.
    \end{aligned}
    \label{2.3}
  \end{equation}
  
  By Lemma S.6 and Condition 1, under the event \(\eta_6\) we have
  \begin{equation}
    \renewcommand\theequation{S.\arabic{equation}}
    \begin{aligned}
      \max_{k \in S^{c(i)}}\lef \mathcal{G}_3 \rig &=\max_{k \in S^{c(i)}}\lef \frac{1}{\tilde\lambda_n}\cdot\onen\tildePsijT \tildePsino \beta_S^{*(-i)} \rig
                                                     \leq  \frac{1}{\tilde\lambda_n} \max_{k \in S^{c(i)}}\lef \onen\tildePsijT \tildePsino \rig \lef \beta_S^{*(-i)} \rig \\
                                                   &\leq  (pd_n)^{-1}\sqrt{s}C\tilde\lambda_n^{-1} \frac{3(\sqrt{\frac{(s-s_i+1)d_n}{n}}+t_{14} )+3(\sqrt{\frac{s-s_i+1}{p}}+t_{15})}{1-3(\sqrt{\frac{s-s_i+{15}}{p}}+t_{16})}.
    \end{aligned}
    \label{2.4}
  \end{equation}
  
  Lastly, we need to bound
  \(\max_{k \in S^{c(i)}} \|\mathcal{G}_4\| = \max_{k \in S^{c(i)}}
  \|T' \e\|\), where \(T' = \tildePsijT F/(\tilde\lambda_n n)\). Just as
  before, we start by conditioning on \(X\), so that \(T'\) becomes a deterministic
  \(d_n\) by \(n\) matrix. This implies
  \(T'\e | X \sim \mathcal{N}(0, \sigma^2T'{T'}^{\T})\), where
  \begin{equation*}
    \begin{aligned}
      \|\sigma^2T'{T'}^{\T}\|&=\frac{\sigma^2}{n^2\tilde\lambda_n^2}\|\tildePsijT FF^{\T}\tildePsij\|\\
                         &\leq \frac{\sigma^2}{n^2\tilde\lambda_n^2} \|F\|^2\left\|\tildePsij\right\|^2.
    \end{aligned}
  \end{equation*}
  As a result,
  \(n\tilde\lambda_n T' \e/(\sigma\|F\|\|\tildePsij\|)\Big|X\) follows
  an \(d_n\)-dimensional Gaussian distribution whose covariance matrix
  has eigenvalues less then \(1\), so we can apply again Lemma 1 of
  \cite{laurent2000adaptive} and taking expectation with respect to
  \(X\),
  \[
    \rP\left(n\tilde\lambda_n \|T' \e\|
      \Big/\left(\sigma\|F\|\|\tildePsij\|\right) \leq
      \sqrt{5sd_n}\right) \geq 1 - \exp(-sd_n).
  \]
  Denoting the event
  \[
    \eta_7 = \{\max_{k \in S^{c(i)}} n\tilde\lambda_n \|T' \e\|
    /(\sigma\|F\|\|\tildePsij\|) \leq \sqrt{5sd_n}\},
  \]
  we have
  \(\rP(\eta_7) \geq 1 - (p_i - s_i)\exp(-sd_n)\) by the union bound.

  Now, under the events \(\eta_5\), \(\eta_7\) and \(\eta_8\), for any
  \(k \in S^{c(i)}\),
  \begin{equation*}
    \begin{aligned}
      \|T' \e\|& \leq \sqrt{5} \sigma n^{-1/2} \tilde\lambda_n^{-1} \sqrt{sd_n} \|F\|\left\|\frac{\tildePsij}{\sqrt{n}}\right\|\\
               & \leq \sqrt{5} \sigma \delta^{-1/2} (p d_n)^{-1} \tilde\lambda_n^{-1} \sqrt{\frac{sd_n}{n^{\alpha}}}\left[1+\frac{3(\sqrt{\frac{d_n}{n}}+t_{11} )+3(\sqrt{\frac{1}{p}}+t_{12})}{1-3(\sqrt{\frac{1}{p}}+t_{13} )}\right]^{\frac{1}{2}}, 
    \end{aligned}
  \end{equation*}
  which in turn implies
  \begin{equation}
    \renewcommand\theequation{S.\arabic{equation}}
    \max_{k \in S^{c(i)}} \|\mathcal{G}_4\| \leq \sqrt{5} \sigma \delta^{-1/2} (p d_n)^{-1} \tilde\lambda_n^{-1} \sqrt{\frac{sd_n}{n^{\alpha}}}\left[1+\frac{3(\sqrt{\frac{d_n}{n}}+t_{11} )+3(\sqrt{\frac{1}{p}}+t_{12})}{1-3(\sqrt{\frac{1}{p}}+t_{13} )}\right]^{\frac{1}{2}}.
    \label{2.5}
  \end{equation}

  Combining the results (S.\ref{2.2}) (S.\ref{2.3}) (S.\ref{2.4})
  (S.\ref{2.5}) with (S.\ref{2.1}) and setting
  $t_{8}=\frac{1}{60\sqrt{s}}-\sqrt{\frac{(s_i+1)d_n}{n}}$,
  $t_{9}=\frac{1}{60\sqrt{s}}-\sqrt{\frac{s_i+1}{p}}$,
  $t_{10}=\frac{1}{10}-\sqrt{\frac{s_i+1}{p}}$,
  $t_{11}=\frac{1}{60}-\sqrt{\frac{d_n}{n}}$,
  $t_{12}=\frac{1}{60}-\sqrt{\frac{1}{p}}$,
  $t_{13}=\frac{1}{10}-\sqrt{\frac{1}{p}}$,
  $t_{14}=\frac{\tilde\lambda_n}{60\sqrt{s}C}-\sqrt{\frac{(s-s_i+1)d_n}{n}}$,
  $t_{15}=\frac{\tilde\lambda_n}{60\sqrt{s}C}-\sqrt{\frac{s-s_i+1}{p}}$,
  $t_{16}=\frac{1}{10}-\sqrt{\frac{s-s_i+1}{p}}$, we have
  \begin{equation*}
    \begin{aligned}
      \max_{k\in S^{c(i)}}\lef pd_n\hatgji \rig &\leq \frac{1}{\tilde\lambda_n}\cdot \frac{3(\sqrt{\frac{(s_i+1)d_n}{n}}+t_8
                                                  )+3(\sqrt{\frac{s_i+1}{p}}+t_9)}{1-3(\sqrt{\frac{s_i+1}{p}}+t_{10}) }\\
                                                &\cdot \left(\frac{\tilde\lambda_n}{6}+\sqrt{\frac{70}{9}}\sigma\delta^{-1/2}n^{-\alpha/2}\sqrt{sd_n}+\sqrt{\frac{7}{6}}C'\delta^{-1/2}n^{(1 - \alpha)/2}sd_n^{-3/2}+\sqrt{\frac{14}{9}}\sqrt{s}\tilde\lambda_n\right)\\
                                                &+C'\delta^{-1/2}\tilde\lambda_n^{-1}n^{(1-\alpha)/2}sd_n^{-3/2}
                                                  \left[1+\frac{3(\sqrt{\frac{d_n}{n}}+t_{11} )+3(\sqrt{\frac{1}{p}}+t_{12})}{1-3(\sqrt{\frac{1}{p}}+t_{13} )}\right]^{\frac{1}{2}}\\
                                                &+\sqrt{s}C\tilde\lambda_n^{-1} \frac{3(\sqrt{\frac{(s-s_i+1)d_n}{n}}+t_{14} )+3(\sqrt{\frac{s-s_i+1}{p}}+t_{15})}{1-3(\sqrt{\frac{s-s_i+1}{p}}+t_{16})}\\
                                                &+\sqrt{5} \sigma \delta^{-1/2} \tilde\lambda_n^{-1} \sqrt{\frac{sd_n}{n^{\alpha}}}\left[1+\frac{3(\sqrt{\frac{d_n}{n}}+t_{11} )+3(\sqrt{\frac{1}{p}}+t_{12})}{1-3(\sqrt{\frac{1}{p}}+t_{13} )}\right]^{\frac{1}{2}}\\
      \leq & \frac{1}{42}s^{-1/2}+\sqrt{\frac{10}{63}}\sigma\delta^{-1/2}\tilde\lambda_n^{-1}n^{-\alpha/2}d_n^{1/2}+\sqrt{\frac{1}{42}}C'\delta^{-1/2}\tilde \lambda_n^{-1}n^{(1 - \alpha)/2}s^{-1/2}d_n^{-3/2}+\sqrt{\frac{2}{63}}\\
                                                &+\sqrt{\frac{8}{7}}C'\delta^{-1/2}\tilde\lambda_n^{-1}n^{(1-\alpha)/2}sd_n^{-3/2} + \frac{1}{7} + \sqrt{\frac{40}{7}}\sigma \delta^{-1/2} \tilde\lambda_n^{-1} n^{-\alpha/2}\sqrt{sd_n}\\
      =& o(1) + \frac{1}{42} + \sqrt{\frac{2}{63}} + \frac{1}{7}
    \end{aligned}
  \end{equation*}
  under events $\eta_1, \eta_2, \dots, \eta_8$ for sufficiently large
  \(n\).

  By Lemma S.5, we have
  \[
    \sqrt{\max_{k\in S^{c(i)}}\lef \left(\frac{pd_n}{n}
        \widetilde{\Psi}_k^{(i)\T}\widetilde{\Psi}_k^{(i)}\right)^{-1}\rig}
    \leq \left[1 - \frac{3(\sqrt{\frac{d_n}{n}}+t_{11}
        )+3(\sqrt{\frac{1}{p}}+t_{12})}{1-3(\sqrt{\frac{1}{p}}+t_{13}
        )}\right]^{-\frac{1}{2}} \leq \sqrt{\frac{7}{6}}
  \]
  under the event \(\eta_5\) for sufficiently large \(n\). This along
  with the previous inequality proves (S.\ref{subgra_bound}).

  Considering the probability for the intersection of $\eta_1, \dots, \eta_8$, we can conclude that
  the solution is sparsistent, i.e. $\hat{S}^{(i)}=S^{(i)}$ with probability at least
  \[1-\exp(-\xi n^\gamma) - 16(p_i - s_i + 1)\exp(-sd_n)\] for
  sufficiently large \(n\). Here we have combined all the exponential
  terms that converge to 0 faster than \(\exp(-sd_n)\) into a single
  term \(16(p_i - s_i + 1)\exp(-sd_n)\).
  
  This completes the proof of Theorem 1.
\end{myproof}

\begin{myproof}[Proof of Corollary 1]
  Note that the definitions of events \(\eta_1\), \(\eta_2\),
  \(\eta_3\) and \(\eta_8\) are independent of the machine index
  \(i\). When considering the probability for the intersection of
  $\eta_1, \dots, \eta_8$ across all machines, those events only need
  to be counted once. Thus, we have \(\hat S = S\) with probability at
  least
  \begin{align*}
    1 &- \exp(-\xi n^\gamma) - 11\exp(-sd_n) - \sum_{i = 1}^m16(p_i - s_i)\exp(-sd_n)\\
    \geq & 1-\exp(-\xi n^\gamma) - 16(p - s + 1)\exp(-sd_n).
  \end{align*}
\end{myproof}

\begin{myproof}[Proof of Theorem 2]
  Here we only prove the case where \(k \in S^{(i)}\). The same argument can
  be easily adapted to the case where \(k \in S^{c(i)}\).

  By the proof of Theorem 1 and Corollary 1, with probability at least
  \(1-\exp(-\xi n^\gamma) - 16(p - s + 1)\exp(-sd_n) \to 1\), we have
  \[
    \left\|\hat\beta_{S^c}\right\| = 0,
  \]
  and
  \begin{equation}
    \renewcommand\theequation{S.\arabic{equation}}
    \label{3.1}
    \begin{aligned}
      &n^{1/4}\left\|\hat\beta_{S}-\beta^{*}_{S}\right\| \\
      &\leq \sqrt{m} n^{1/4}\left(\frac{\tilde\lambda_n}{6}+\sqrt{\frac{70}{9}}\sigma\delta^{-1/2}n^{-\alpha/2}\sqrt{sd_n}+\sqrt{\frac{7}{6}}C'\delta^{-1/2}n^{(1 - \alpha)/2}sd_n^{-3/2}+\sqrt{\frac{14}{9}}\sqrt{s}\tilde\lambda_n\right)\\
      &\lesssim \sqrt{s} \tilde \lambda_n n^{1/4}(s^{-1/2} + \tilde \lambda_n^{-1}n^{-\alpha/2}\sqrt{d_n} + \tilde \lambda_n^{-1}n^{(1-\alpha)/2}s^{1/2}d_n^{-3/2} + 1) \to 0.
    \end{aligned}
  \end{equation}
  The centerpiece of our proof is the following decomposition,
  \begin{align*}
    \widehat{M}^{(i)}_k \left(\hat\beta^{u(i)}_k - \beta^{*(i)}_k\right) = & \widehat{M}^{(i)}_k \left(\hat\beta^{(i)}_k - \beta^{*(i)}_k + \frac{pd_n}{n}
                                                           \tildePsijT \left(\widetilde Y - \widetilde\Psi\hat\beta\right)\right)\\
    = & \widehat{M}^{(i)}_k \left(\hat\beta^{(i)}_k - \beta^{*(i)}_k\right)\\
                                                         &+\frac{pd_n}{n} \widehat{M}^{(i)}_k \tildePsijT \left(\tildePsi \beta^{*(i)}_S + \tildePsino \beta^{*(-i)}_S + \widetilde Z + \tilde\e - \tildePsi \hat\beta^{(i)}_S - \tildePsino \hat\beta^{(-i)}_S\right)\\
    = & \widehat{M}^{(i)}_k \left(I^{(i)}_k - \frac{pd_n}{n}\tildePsijT \tildePsi \right)\left(\hat\beta^{(i)}_S - \beta^{*(i)}_S\right)\\
                                                         &+ \frac{pd_n}{n} \widehat{M}^{(i)}_k \tildePsijT \tildePsino\left(\beta^{*(-i)}_S - \hat\beta^{(-i)}_S\right) +\frac{pd_n}{n} \widehat{M}^{(i)}_k \tildePsijT \widetilde Z + \frac{pd_n}{n} \widehat{M}^{(i)}_k \tildePsijT \tilde \e \\
    : = & \Delta_1 + \Delta_2 + \Delta_3 + w, 
  \end{align*}
  where \(I^{(i)}_k\) is a \(d_n\) by \(s_id_n\) matrix, whose entries
  are all zeroes except for an identity submatrix \(I_{d_n}\) on
  columns \(kd_n + 1\) through \((k + 1)d_n\).

  First, we have \(w = \widetilde T \e\), where
  \(\widetilde T = pd_n \widehat{M}^{(i)}_k \tildePsijT F/n\). Conditioned on \(X\),
  \(w\) follows a mean zero normal distribution with covariance
  \(\sigma^2\widetilde T \widetilde T^{\T} = \sigma^2p^2d_n^2\widehat{M}^{(i)}_k
  \tildePsijT F F^{\T} \tildePsij \widehat{M}^{(i)\T}_k/n^2= I_{d_n}\).  Taking
  expectation with respect to \(X\), we get \(w \sim \mathcal{N}(0, I_{d_n})\).
  
  Next, we will show that
  \(\|\Delta\| : = \|\Delta_1 + \Delta_2 + \Delta_3\| \leq
  \|\Delta_1\| + \|\Delta_2\| + \|\Delta_3\| \to 0\) under additional
  events \(\eta_1\) \(\eta_2\) and \(\eta_9\) (as defined below). For
  \(\eta_1\) and \(\eta_2\), we choose
  \(t_1 = t_4 = \frac{1}{60n^{1/4}}-\sqrt{\frac{sd_n}{n}}\),
  \(t_2 = t_5 = \frac{1}{60n^{1/4}}-\sqrt{\frac{s}{p}}\) and
  \(t_3 = t_6 = \frac{1}{10}-\sqrt{\frac{s}{p}}\), such that
  \(\rP(\eta_1) \wedge \rP(\eta_2) \geq 1 - 5\exp(-n^{1/2}/28800)\)
  and they imply
  \begin{equation}
    \renewcommand\theequation{S.\arabic{equation}}
    \label{3.2}
    \left\|I^{(i)}_k - \frac{pd_n}{n}\tildePsijT \tildePsi \right\|
    \leq \frac{1}{7n^{1/4}}, 
  \end{equation}
  \[
    \left\|\frac{pd_n}{n}\tildePsijT \tildePsino\right\|\leq \frac{1}{7n^{1/4}}, 
  \]
  and
  \[
    \left\|\left(\frac{pd_n}{n}\tildePsijT
        \tildePsij\right)^{-1}\right\| \leq \frac{7}{6}.
  \]
  
  By Condition 6, we get
  \[
    \rP\left(\|F^{-1}\| \leq \sqrt{pd_n\delta'}\right) \geq 1 - \exp(-\xi' n^{\gamma'}).
  \]
  Denote the smallest singular value of a matrix by
  \(\sigma_{\min}\). Under the events
  \[
    \eta_9: = \{\|F^{-1}\| \leq \sqrt{pd_n\delta'}\}
  \]
  and \(\eta_2\),
  \begin{equation}
    \renewcommand\theequation{S.\arabic{equation}}
    \label{3.3}
    \begin{aligned}
      \|\widehat{M}^{(i)}_k\| =& \frac{n}{pd_n\sigma\sqrt{\sigma_{\min}(\tildePsijT F F^{\T} \tildePsij)}}\\
      \leq& \frac{n}{pd_n\sigma\sigma_{\min}(\tildePsij) \sigma_{\min}(F)}\\
      = & \frac{n}{pd_n\sigma} \sqrt{\left\|(\tildePsijT \tildePsij)^{-1}\right\|}\|F^{-1}\|\\
      \leq & \sqrt{\frac{7\delta'}{6\sigma^2}} n^{1/2}.
    \end{aligned}
  \end{equation}
  
  By (S.\refeq{3.1}), (S.\refeq{3.2}) and (S.\refeq{3.3}), we have
  \(\|\Delta_1\| \leq \|\widehat{M}^{(i)}_k\| \|I^{(i)}_k - pd_n\tildePsijT \tildePsi
  /n\| \|\hat\beta_{S}-\beta^{*}_{S}\| \leq
  \sqrt{\delta'/(42\sigma^2)} n^{1/4} \|\hat\beta_{S}-\beta^{*}_{S}\|
  \to 0\).

  Similarly,
  \(\|\Delta_2\| \leq \|\widehat{M}^{(i)}_k\| \|pd_n \tildePsijT \tildePsino
  /n\|\|\hat\beta_{S}-\beta^{*}_{S}\| \leq \sqrt{\delta'/(42\sigma^2)}
  n^{1/4} \|\hat\beta_{S}-\beta^{*}_{S}\| \to 0\)
  
  Since \(\widetilde T \widetilde T^{\T} = I_{d_n}/\sigma^2\), it
  follows \(\|\widetilde T\| = 1/\sigma\). By (S.\refeq{zequation})
  and Condition 7,
  \(\|\Delta_3\| = \|\widetilde T Z\| \leq \frac{\sqrt{n}sC'}{\sigma
    d_n^{3/2}} \to 0\).

  This completes the proof of Theorem 2.
\end{myproof}

\putbib

\end{bibunit}

\end{document}